\documentclass[letterpaper,journal]{IEEEtran}
\usepackage{amsmath,amsfonts}
\usepackage{algorithmic}
\usepackage{algorithm}
\usepackage{array}
\usepackage[caption=false,font=normalsize,labelfont=sf,textfont=sf]{subfig}
\usepackage{textcomp}
\usepackage{stfloats}
\usepackage{url}
\usepackage{capt-of}
\usepackage{verbatim}
\usepackage{graphicx}
\usepackage{float}
\usepackage{cite}
\usepackage[colorlinks,urlcolor=blue,linkcolor=blue,citecolor=blue]{hyperref}
\usepackage{multirow}
\usepackage{xcolor}
\usepackage{array}
\usepackage{pifont}
\usepackage{hyperref}
\usepackage{booktabs}
\newcommand{\xmark}{\ding{55}}
\newcolumntype{C}[1]{>{\centering\arraybackslash}m{#1}}
\begin{document}

% \title{The Way for Xingxing Yang to the Paradise}
% \title{Towards High-Quality Low-Light Remote Sensing Image Enhancement via Semantic and Geometric Prior Guidance}
\title{Overcoming Attention Drift: Homogeneity-Heterogeneity Guided Feature Aggregation for Low-Light Remote Sensing Image Enhancement}

% \title{Breaking Erroneous Aggregation: Semantic Homogeneity and Topological Heterogeneity for Low-Light Remote Sensing Image Enhancement}
% \title{Homogeneity-Heterogeneity Guided Feature Aggregation via Dual Foundation Priors for Low-Light Remote Sensing Image Enhancement}

% \author{Authors
        
% \author{Xingxing Yang,
%         Yaozi Zhong,~\IEEEmembership{Student Member,~IEEE},
%         Shaohui Mei,~\IEEEmembership{Senior Member,~IEEE},
%         and~Mingyang Ma
% % \thanks{This paper was produced by the IEEE Publication Technology Group. They are in Piscataway, NJ.}% <-this % stops a space
% \thanks{Xingxing Yang and Yaozi Zhong contributed equally to this work. Corresponding author: Xingxing Yang, Email:csxxyang@comp.hkbu.edu.hk}
% \thanks{Xingxing Yang is with the Department of Computer Science, Hong Kong Baptist University, Hong Kong, China}
% \thanks{Yaozi Zhong is with the School of Information and Artificial Intelligence, Yunnan University of Finance and Economics, Kunming, China.}
% \thanks{Shaohui Mei and Mingyang Ma are with the School of Electronics and Information, Northwestern Polytechnical University, Xi’an 710129, China}
% }

\author{Yaozi Zhong,
        Xingxing Yang,
        Shaohui Mei,~\IEEEmembership{Senior Member,~IEEE},
        and~Mingyang Ma
\thanks{Yaozi Zhong and Xingxing Yang contributed equally to this work. 
Corresponding author: Xingxing Yang, Email: csxxyang@comp.hkbu.edu.hk}
\thanks{Yaozi Zhong is with the School of Information and Artificial Intelligence, Yunnan University of Finance and Economics, Kunming, China.}
\thanks{Xingxing Yang is with the Department of Computer Science, Hong Kong Baptist University, Hong Kong, China.}
\thanks{Shaohui Mei and Mingyang Ma are with the School of Electronics and Information, Northwestern Polytechnical University, Xi'an 710129, China.}
}
% \markboth{IEEE Trans. Geosci. Remote Sens.}%
% {Yang \MakeLowercase{\textit{et al.}}: Overcoming Attention Drift}

% \makeatletter
% \let\@oldmaketitle\@maketitle
% \renewcommand{\@maketitle}{\@oldmaketitle
% \vspace{-1em}
% \begin{center}
% \includegraphics[width=\textwidth]{figures/fig1.pdf}
% \vspace{-0.2cm}
% \captionof{figure}{Visual comparison of attention mechanisms and enhancement results. (a) A query token (blue cross) on a tent edge. (b) LLFormer~\cite{Wang2023Ultra} suffers from spatial propagation errors, leading to Type I (Cross-boundary Confusion) and Type II (Coplanar Ambiguity) drifts. (c) With dual priors, HALO constrains attention within the target, enabling coherent aggregation and sharp boundary preservation. (d) Due to attention drift, LLFormer exhibits blurred edges and color distortion, while HALO produces sharper and more color-faithful results.}
% \label{fig:comparison}
% \end{center}
% \vspace{1.5em}
% }
% \makeatother

\makeatletter
\let\@oldmaketitle\@maketitle
\renewcommand{\@maketitle}{\@oldmaketitle
% \vspace{-1em}
% \begin{center}
% \includegraphics[width=\textwidth]{figures/fig1-crop.pdf}
% \vspace{-0.3cm}
% \setcounter{figure}{0}
% \captionof{figure}{Visual comparison of attention mechanisms and enhancement results. (a) A query token (blue cross) on a tent edge. (b) LLFormer~\cite{Wang2023Ultra} suffers from spatial propagation errors, leading to Type I (Cross-boundary Confusion) and Type II (Coplanar Ambiguity) drifts. (c) With dual priors, HALO constrains attention within the target, enabling coherent aggregation and sharp boundary preservation. (d) Due to attention drift, LLFormer exhibits blurred edges and color distortion, while HALO produces sharper and more color-faithful results.}
% \label{fig:teaser}
% \end{center}
\vspace{-0.5em}
}
\makeatother

\maketitle

\begin{abstract}

Restoring high-fidelity remote sensing imagery from extreme low-light degradation is indispensable for reliable Earth observation and downstream machine vision. However, under severe noise and illumination corruption, existing methods suffer from attention drift, erroneously aggregating features across distinct physical boundaries and causing severe structural blurring and color distortion. To address this, we propose HALO, a dual-prior-driven enhancement framework that formulates enhancement as a guided feature aggregation problem driven by foundation model priors. Specifically, an illumination-invariant semantic prior provides regional homogeneity as a positive bias for content-consistent aggregation, while a pseudo-3D topological prior provides boundary heterogeneity as a negative penalty to strictly prevent cross-boundary confusion. To cooperatively incorporate these two priors, we propose a Homogeneity-Heterogeneity Cooperative Attention Module (H2CAM) to resolve feature conflicts during cross-modal prior fusion. Extensive experiments demonstrate that HALO achieves state-of-the-art performance across 8 challenging synthetic and real-world remote sensing benchmarks, significantly improving physical boundary sharpness and color fidelity while maximizing the preservation of discriminative features. Code is publicly available at: \href{https://github.com/AlexYangxx/HALO}{\textcolor{cyan}{HALO}}.

\end{abstract}


\section{Introduction}

\IEEEPARstart{E}{arth} observation heavily relies on clear remote sensing imagery to perform reliable downstream machine vision tasks such as object detection and land cover classification~\cite{Cheng2017Remote,Zhu2017Deep,xia2018dota}. Extreme low-light remote sensing is not simply a matter of enhancing visual brightness. Instead, the fundamental objective is to strictly preserve physical boundaries and color fidelity. Complex remote sensing scenes often feature long-range continuous surface structures and heterogeneous materials~\cite{Zhao2025Atmospheric}, making this preservation exceptionally challenging under severe illumination degradation.

\begin{figure}[!t]
\centering
\includegraphics[width=0.4\textwidth]{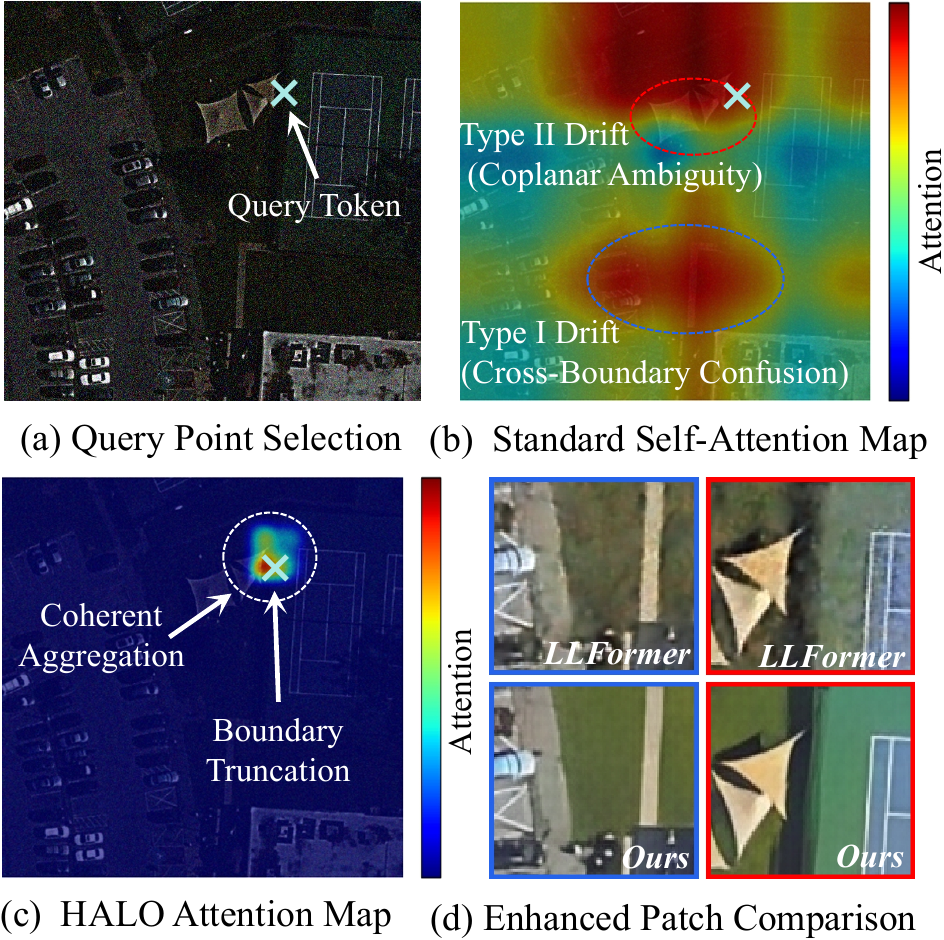}
\captionof{figure}{Motivation illustration. (a) A query token (blue cross) on a tent edge. (b) LLFormer~\cite{Wang2023Ultra} suffers from spatial propagation errors, leading to Type I (Cross-boundary Confusion) and Type II (Coplanar Ambiguity) drifts. (c) With dual priors, HALO constrains attention within the target, enabling coherent aggregation and sharp boundary preservation. (d) Due to attention drift, LLFormer exhibits blurred edges and color distortion, while HALO produces sharper and more color-faithful results.}
\label{fig:teaser}
\vspace{-5pt}
\end{figure}

% Existing paradigms primarily rely on the internal statistics of degraded observations~\cite{Chen2018Retinex,Wang2023FourLLIE,Yao2024Spatial}. Under extreme degradation where the signal-to-noise ratio approaches zero, this pure data-driven approach completely collapses~\cite{Li2022Low,Ma2022Toward}. Recent methods~\cite{Wang2022Uformer, Yan2025HVI} depend on self-attention mechanisms to capture global dependencies. However, recent studies observe that these unconstrained interactions often fail in extremely dark regions, leading to severe noise amplification and structural distortion~\cite{yin2025Structure,Li2025SAIGFormer}. \textit{We formalize this corrupted internal statistic fitting as a mathematical failure, which we define as \textbf{Attention Drift}}. This drift manifests as two fatal physical crises, as illustrated in Fig.~\ref{fig:teaser}. First, noise-induced high-frequency responses cause physically disjoint tokens to erroneously aggregate into similar low-light responses, resulting in blurred structural truncations and creating a Type I Error: \textbf{\textit{Cross-Boundary Confusion}}. Second, severe illumination variations fragment continuous homogeneous materials into disconnected clusters, causing color distortion and semantic leakage, identified as Type II Error: \textbf{\textit{Coplanar Ambiguity}}.

Existing paradigms primarily rely on the internal statistics of degraded observations~\cite{Chen2018Retinex,Wang2023FourLLIE,Yao2024Spatial}. Under extreme degradation where the signal-to-noise ratio approaches zero, this pure data-driven approach completely collapses~\cite{Li2022Low,Ma2022Toward}. Recent methods~\cite{Wang2022Uformer, Yan2025HVI} depend on self-attention mechanisms to capture global dependencies. However, recent studies observe that these unconstrained interactions often fail in extremely dark regions, leading to severe noise amplification and structural distortion~\cite{yin2025Structure,Li2025SAIGFormer}. \textit{We formalize this corrupted internal statistic fitting as a mathematical failure, which we define as \textbf{Attention Drift}}. This drift manifests as two fatal physical crises, as illustrated in Fig.~\ref{fig:teaser}. First, noise-induced high-frequency responses cause physically disjoint tokens to yield spuriously high affinities, blurring structural truncations and creating a Type I Error: \textbf{\textit{Cross-Boundary Confusion}}. Second, extreme illumination degradation forces distinct physical materials to collapse into identical dark responses while fragmenting continuous homogeneous regions. This causes severe semantic leakage and color distortion, identified as Type II Error: \textbf{\textit{Coplanar Ambiguity}}.

\begin{figure*}[!t]
\centering
\includegraphics[width=0.9\textwidth]{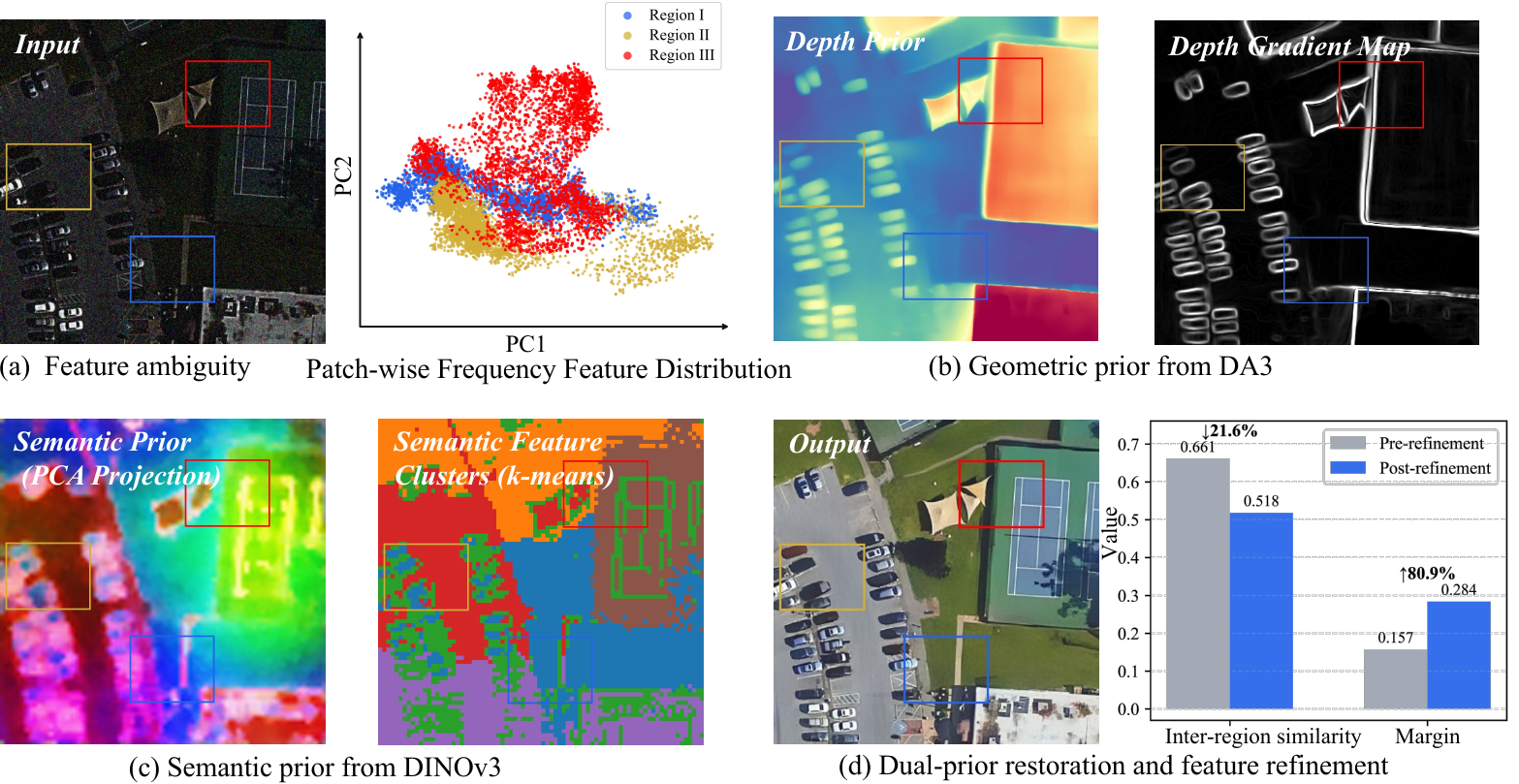}
\caption{Performance illustration. (a) Low-light degradation causes feature entanglement. (b) and (c) Geometric and semantic priors provide reliable structural and semantic cues. (d) Integrating them reduces erroneous similarity by \textbf{21.6\%} and boosting the discriminative margin by \textbf{80.9\%}.}
\vspace{-5pt}
\label{fig:defect_analysis}
\end{figure*}

\IEEEpubidadjcol
To this end, relying solely on pure internal statistical fitting is often insufficient, which motivates us to incorporate external deterministic physical constraints. Following the recent trend of utilizing robust foundation models for complex visual restoration~\cite{Lin2026PixIE}, we propose HALO, driven by dual foundation model priors for low-light remote sensing image enhancement.
As shown in Fig.~\ref{fig:defect_analysis}, we leverage an illumination-invariant semantic prior from DINOv3~\cite{Simeoni2025DINOv3} to provide regional homogeneity. This acts as a positive homogeneity bias to force content-consistent aggregation, directly solving the Type II Error~\cite{Wu2023Learning,Zheng2022Semantic}. Simultaneously, we extract a pseudo-3D topological prior from Depth Anything 3~\cite{Lin2026Depth} to provide boundary heterogeneity. This operates as a negative heterogeneity penalty to strictly cut off feature propagation across structural edges, completely resolving the Type I Error~\cite{Wang2024Depth}.
To effectively incorporate these two priors, we design the Homogeneity-Heterogeneity Cooperative Attention Module to translate these foundation priors into explicit mathematical boundaries within the attention matrix, replacing unconstrained data fitting with a bounded physical optimization. Extensive evaluations demonstrate that HALO yields substantial performance gains for downstream Earth observation object detection tasks, decisively outperforming existing methods in both physical boundary sharpness and color fidelity.

In summary, our main contributions are as follows:
\begin{itemize}
\item \textbf{Theoretical Insight}: We reveal the theoretical bottleneck of Attention Drift and demonstrate the absolute necessity of introducing deterministic external physical constraints for low-light remote sensing enhancement.
\item \textbf{Framework Innovation}: We propose HALO, a new dual-prior framework that formulates image enhancement as a guided feature aggregation problem to overcome attention failure under extreme degradation.
\item \textbf{Core Mechanism}: We design the Homogeneity-Heterogeneity Cooperative Attention Module to explicitly translate foundation model priors into positive and negative mathematical biases within the attention matrix.
\item \textbf{Empirical Generalization}: HALO achieves superior physical boundary sharpness and color fidelity, demonstrating powerful empirical generalization and delivering significant performance improvements for downstream machine vision tasks.
\end{itemize}

\section{Related Work}

\subsection{Low-Light Remote Sensing Image Enhancement}

Existing low-light remote sensing enhancement methods primarily evolve from localized spatial restoration to global context modeling. Early spatial paradigms, including Retinex-based CNNs~\cite{Chen2018Retinex,Guo2020ZeroDCE} and recent generative diffusion frameworks~\cite{QWR2026,ECADiff2025}, establish robust foundations for illumination correction and texture synthesis. However, their restoration trajectory heavily depends on the internal statistics of degraded observations. Because heterogeneous land covers often exhibit identical dark responses in remote sensing scenes, these pure data-driven methods struggle to distinguish underlying physical properties, frequently suffering from coplanar ambiguity and semantic fragmentation.

To capture the long-range continuous structures essential for Earth observation, Transformer architectures~\cite{Wang2022Uformer} and joint spatial frequency networks~\cite{Wang2023FourLLIE,Zhang2026SPJFNet} have become the dominant paradigms. While these architectures successfully expand receptive fields and improve computational efficiency, a fundamental structural gap remains. Under severe noise and extreme illumination variation, the attention affinities computed directly from corrupted features degenerate, triggering attention drift. Furthermore, because structural boundaries and high-frequency noise are densely superimposed in the frequency domain, these unconstrained architectures lack deterministic criteria for feature aggregation, inevitably leading to erroneous cross-region propagation and structural blurring.

\subsection{Multimodal Prior Mechanisms in Visual Restoration}

External priors are indispensable for compensating for severe information loss in degraded observations~\cite{Awais2025Foundational}. The field has progressively transitioned from handcrafted physical priors, such as gradient and illumination maps~\cite{Nazeri2019EdgeConnect,Wang2024Depth}, to robust foundation model representations. Recent architectures leverage multi-modal large models~\cite{LMDIR2024} and diffusion priors~\cite{Tang2025DSPFusion,TPGDiff2026} for scene-aware contextual guidance. In this context, DINOv3~\cite{Simeoni2025DINOv3} extracts illumination-invariant semantic features for region-level consistency, while DA3~\cite{Lin2026Depth} estimates pseudo-3D topological cues for robust boundary modeling.

Nevertheless, a critical methodological limitation persists regarding how these external priors are integrated. Current prior-guided networks predominantly employ feature concatenation, cross-attention, or spatial modulation~\cite{Yu2025Multiprior}. These standard integration strategies treat external priors merely as auxiliary input data rather than explicit attention constraints. Consequently, when optimizing under extremely low-light degradation, the network optimization remains dominated by the corrupted observations. The incorporated foundation priors are rapidly diluted and assimilated by noisy features during subsequent unconstrained aggregation operations. This soft integration renders them powerless to fundamentally prevent semantic confusion, necessitating a paradigm shift toward deterministic prior-constrained attention mechanisms.

\section{Proposed Method}
\subsection{Problem Formulation: Attention Drift under Extreme Degradation}
Low-light remote sensing enhancement aims to recover a clear image $J \in \mathbb{R}^{H \times W \times 3}$ from a degraded observation
\begin{equation}
I(x) = J(x) \odot L(x) + N(x),
\end{equation}
where $L(x)$ is the non-uniform illumination and $N(x)$ is heavy noise. The near-zero signal-to-noise ratio severely corrupts the internal statistics of $I(x)$.

Previous frameworks~\cite{Yao2024Spatial, Li2025LersGAN} typically employ Self-Attention for feature aggregation. For a given feature set $Z$, the unconstrained affinity matrix is computed as $A = \mathrm{Softmax}(QK^\top / \sqrt{d})$. Ideally, the attention weight $A_{i,j} \propto \exp(q_i k_j^\top)$ should reflect the true physical correlation between token $i$ and token $j$. However, under extreme degradation, the inner product is dominated by degradation terms rather than the underlying signal $J$. This leads to a fatal phenomenon we define as \textbf{Attention Drift} (visually demonstrated in Fig.~\ref{fig:teaser}), mathematically approximated as:
\begin{equation}
    A_{i,j}^{\text{degraded}} \approx \mathrm{Softmax} \left( \frac{(q_i^J \odot l_i + n_i)(k_j^J \odot l_j + n_j)^\top}{\sqrt{d}} \right).
\end{equation}
As the illumination intensity $l \to 0$ and noise $n$ intensifies, the true semantic correlation $\langle q_i^J, k_j^J \rangle$ is overwhelmed by two critical errors:
\begin{itemize}
    \item \textbf{Type I: Cross-boundary Confusion.} Spurious high-frequency noise responses $\langle n_i, n_j \rangle$ cause physically disjoint tokens to erroneously yield high affinities, crossing physical edges and causing structural blurring.
    \item \textbf{Type II: Coplanar Ambiguity.} Extreme illumination degradation $\langle l_i, l_j \rangle$ forces distinct physical materials to collapse into identical dark responses, while highly non-uniform lighting fragments continuous homogeneous regions. This assigns them erroneous or mixed restoration strategies, causing severe semantic leakage and spectral shifts.
\end{itemize}

\textbf{Prior-Guided Deterministic Aggregation.} Relying solely on internally degraded statistics is theoretically ill-posed. To overcome \textit{Attention Drift}, we transform the unconstrained attention into a deterministic prior-guided aggregation process. We inject an illumination-invariant semantic prior $F_{\mathrm{sem}}$ and a pseudo-3D topological prior $F_{\mathrm{geo}}$ to bound the attention affinity:
\begin{equation}
\hat{A}_{i,j} = \mathrm{Softmax} \left( \frac{q_i k_j^\top}{\sqrt{d}} + \underbrace{\lambda_s \mathcal{H}(F_{\mathrm{sem}})_{i,j}}_{\text{Homogeneity Bias}} + \underbrace{\lambda_g \mathcal{P}(F_{\mathrm{geo}})_{i,j}}_{\text{Heterogeneity Penalty}} \right),
\end{equation}
where the homogeneous modulation $\mathcal{H}(\cdot)$ acts as a positive bias enforcing content-consistent aggregation within semantic regions (resolving Type II error), and the heterogeneous modulation $\mathcal{P}(\cdot)$ acts as a negative penalty strictly halting feature propagation across physical edges (resolving Type I error). This guarantees a physically compliant restoration trajectory $\hat{J} = \mathcal{N}_{\mathrm{HALO}}(I; \hat{A})$.

\begin{figure*}[!t]
\centering
\includegraphics[width=0.95\textwidth]{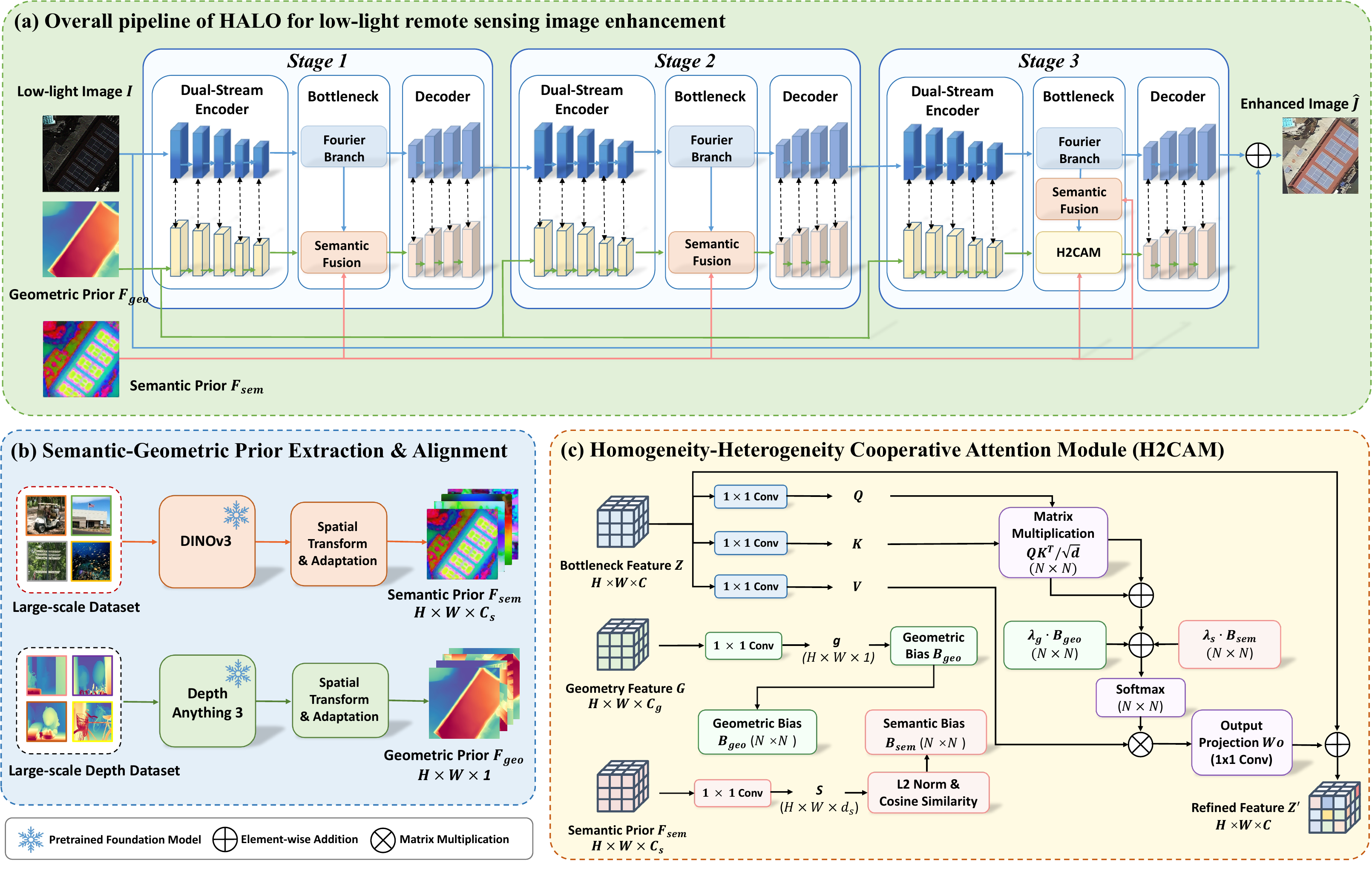}
\caption{Overview of the HALO framework. (a) A cascaded pipeline restores features using dual-stream encoding and decoding. (b) Semantic and geometric priors are extracted via DINOv3 and DA3, then aligned to the restoration space. (c) At the bottleneck, our Homogeneity-Heterogeneity Cooperative Attention Module (H2CAM) integrates these priors as deterministic biases to explicitly suppress cross-boundary leakage and enforce regional consistency.}

\label{fig:pipeline}
\end{figure*}

\subsection{Dual Prior Driven Feature Aggregation Framework}
To implement this prior-guided aggregation, we adopt a cascaded spatial frequency architecture to efficiently process the high-resolution and long-range continuous structures typical of remote sensing images, as shown in Fig.~\ref{fig:pipeline}(a). Rather than treating the external priors as simple concatenated auxiliary inputs, we elevate them into explicit guiding streams that modulate the entire restoration trajectory. 

Given a degraded low-light image $I$, an aligned illumination invariant semantic prior $F_{\mathrm{sem}}$, and an aligned pseudo-3D topological prior $F_{\mathrm{geo}}$, the network first projects the input into the restoration feature space and embeds the topological prior into a dedicated geometry-aware stream:
\begin{equation}
F_0=\phi_{\mathrm{in}}(I), \qquad G_0=\phi_{\mathrm{geo}}(F_{\mathrm{geo}}),
\end{equation}
where $F_0$ denotes the initial image feature, and $G_0$ represents the topological boundary descriptor.

The framework consists of three cascaded encoder-bottleneck-decoder stages. Within this architecture, the dual priors serve distinct physical roles. The topological prior $G_0$ interacts with the image features throughout all stages, acting as a persistent structural anchor to prevent high-frequency noise from spreading across physical truncations. Conversely, the semantic prior $F_{\mathrm{sem}}$ is injected specifically at the bottleneck of each stage, where the receptive field is maximized, serving as a region-level homogeneity anchor to correct semantic leakage across large homogeneous areas.

Let $F_t$ denote the output feature of the $t$ th restoration stage. The cascaded guided aggregation is formulated as
\begin{equation}
F_t =
\begin{cases}
\mathcal{F}_t(F_{t-1}, G_0, F_{\mathrm{sem}}), & t=1,2, \\
\mathcal{F}_3^{\star}(F_2, G_0, F_{\mathrm{sem}}), & t=3,
\end{cases}
\end{equation}
where $\mathcal{F}_t(\cdot)$ denotes the standard prior-guided restoration stage, and $\mathcal{F}_3^{\star}(\cdot)$ incorporates the ultimate Homogeneity-Heterogeneity Cooperative Attention Module (H2CAM) at its bottleneck to finalize the deterministic feature aggregation. The enhanced image is generated through global residual reconstruction: $\hat{J}=\phi_{\mathrm{out}}(F_3)+I$.

\subsection{Semantic Geometric Prior Extraction and Alignment}
As illustrated in Fig.~\ref{fig:pipeline}(b), HALO uses frozen vision foundation models to extract external priors that remain robust under severe illumination degradation. Given the preprocessed inputs $I_{\mathrm{sem}}$ and $I_{\mathrm{geo}}$, the raw priors are obtained as
\begin{equation}
F_{\mathrm{sem}}^{0}=\Phi_{\mathrm{sem}}(I_{\mathrm{sem}}), \qquad F_{\mathrm{geo}}^{0}=\Phi_{\mathrm{geo}}(I_{\mathrm{geo}}),
\end{equation}
where $\Phi_{\mathrm{sem}}(\cdot)$ denotes the DINOv3~\cite{Simeoni2025DINOv3} extractor yielding a dense semantic feature map, and $\Phi_{\mathrm{geo}}(\cdot)$ denotes the Depth Anything 3~\cite{Lin2026Depth} estimator providing a pseudo-3D depth map.

To ensure spatial consistency with the restoration input during training, the image, semantic prior, and geometric prior share the exact same spatial transform $\mathcal{T}(\cdot)$ (e.g., random cropping and flipping). For the semantic prior, the crop window is specifically mapped from the image space to the corresponding token grid. After alignment and interface adaptation, the priors are formulated as
\begin{equation}
F_{\mathrm{geo}}=\mathcal{T}(F_{\mathrm{geo}}^{0}), \qquad F_{\mathrm{sem}}=\mathcal{A}_{\mathrm{sem}}(\mathcal{T}(F_{\mathrm{sem}}^{0})),
\end{equation}
where $\mathcal{A}_{\mathrm{sem}}(\cdot)$ denotes the semantic adaptation operator, including resolution matching and channel projection. These aligned priors subsequently drive the guided aggregation process across all restoration stages. The reliability of the DA3-derived geometric prior under severe low-light degradation is further validated in \textcolor{blue}{Appendix Sec.~IV} through consistency and boundary-overlap analysis.

\subsection{Homogeneity-Heterogeneity Cooperative Attention Module}
As discussed in our problem formulation, the standard self-attention mechanism is highly susceptible to \textit{Attention Drift} under severe low-light conditions. To instantiate the deterministic prior-guided aggregation proposed in Eq. (3), we design the Homogeneity-Heterogeneity Cooperative Attention Module (H2CAM). H2CAM is embedded at the bottleneck of the final restoration stage, where multi-scale contextual features have been aggregated to support semantic and geometric reasoning. 

Unlike previous works~\cite{Wu2023Learning,Lu2025DeepSPG} that merely concatenate prior features or use them for spatial modulation, H2CAM fundamentally reforms the attention affinity matrix by explicitly translating the foundation model priors into mathematical biases and penalties. As illustrated in Fig.~\ref{fig:pipeline}(c), given the normalized bottleneck feature $\bar{Z}=\mathrm{Norm}(Z)$, we generate the query $Q$, key $K$, and value $V$ using learnable $1\times1$ projections, and partition them into local windows (yielding $N$ tokens per window). The unconstrained, data-driven internal affinity within each window is represented by $\frac{QK^\top}{\sqrt{d}}$, which is vulnerable to noise and non-uniform illumination. To correct this, we introduce dual deterministic physical constraints.

\begin{table*}[!t]
\centering
\caption{Quantitative comparisons on full-reference benchmarks, including iSAID-dark~\cite{Yao2024Spatial}, iSAID-dark (high-pixel)~\cite{Yao2024Spatial}, UCM-RSLL, and LOLv1~\cite{Chen2018Retinex}. The best, second-best, and third-best results are highlighted in \textcolor{red}{red}, \textcolor{cyan}{cyan}, and \textcolor{green}{green}, respectively.}
\label{tab:quantitative_comparison_main}
\renewcommand{\arraystretch}{1.1}
\resizebox{\textwidth}{!}{
\begin{tabular}{l|c|ccc|ccc|ccc|ccc}
\noalign{\hrule height 1.2pt}
\multirow{2}{*}{Methods} & \multirow{2}{*}{Venue}
& \multicolumn{3}{c|}{iSAID-dark}
& \multicolumn{3}{c|}{iSAID-dark (high-pixel)}
& \multicolumn{3}{c|}{UCM-RSLL}
& \multicolumn{3}{c}{LOLv1} \\
\cline{3-14}
& & \rule{0pt}{8pt} PSNR$\uparrow$ & SSIM$\uparrow$ & LPIPS$\downarrow$
& PSNR$\uparrow$ & SSIM$\uparrow$ & LPIPS$\downarrow$
& PSNR$\uparrow$ & SSIM$\uparrow$ & LPIPS$\downarrow$
& PSNR$\uparrow$ & SSIM$\uparrow$ & LPIPS$\downarrow$ \\
\hline
% SNR-Aware~\cite{Xu2022SNR} & CVPR'22
% & 23.223 & 0.748 & 0.249
% & \textcolor{cyan}{24.449} & 0.720 & 0.321
% & 19.928 & 0.732 & \textcolor{green}{0.230}
% & \textcolor{red}{24.610} & 0.842 & 0.146 \\
Uformer~\cite{Wang2022Uformer} & CVPR'22
& 24.166 & 0.769 & 0.227
& 16.845 & 0.689 & 0.333
& 18.515 & 0.708 & 0.260
& 18.547 & 0.784 & 0.321 \\
URetinexNet~\cite{Wu2022URetinexNet} & CVPR'22
& 23.494 & 0.740 & 0.276
& 21.203 & 0.652 & 0.383
& 17.884 & 0.539 & 0.482
& 21.328 & 0.826 & 0.238 \\
CUE~\cite{Zheng2023CUE} & ICCV'23
& 21.796 & 0.693 & 0.416
& 20.568 & 0.728 & 0.390
& 19.299 & 0.667 & 0.394
& 21.688 & 0.769 & 0.151 \\
FourLLIE~\cite{Wang2023FourLLIE} & ACM MM'23
& 21.931 & 0.682 & 0.327
& 17.775 & 0.546 & 0.513
& 19.665 & 0.688 & 0.288
& 20.074 & 0.741 & 0.166 \\
LLFormer~\cite{Wang2023Ultra} & AAAI'23
& 24.345 & 0.738 & 0.276
& 23.226 & 0.702 & 0.353
& 19.763 & 0.717 & 0.281
& 23.653 & 0.816 & 0.169 \\
LANet~\cite{Yang2023LANet} & IJCV'23
& 16.899 & 0.421 & 0.695
& 16.770 & 0.479 & 0.684
& 19.009 & 0.559 & 0.788
& 21.745 & 0.816 & 0.101 \\
NeRCo~\cite{Yang2023NeRCo} & ICCV'23
& 21.421 & 0.648 & 0.326
& 19.675 & 0.602 & 0.736
& \textcolor{cyan}{20.584} & 0.713 & 0.403
& 22.946 & 0.801 & 0.311 \\
PairLIE~\cite{Fu2023PairLIE} & CVPR'23
& 15.997 & 0.271 & 0.830
& 14.525 & 0.290 & 0.732
& 17.531 & 0.404 & 0.600
& 19.510 & 0.736 & 0.248 \\
DFFN~\cite{Yao2024Spatial} & TGRS'24
& \textcolor{green}{25.541} & 0.785 & 0.217
& \textcolor{green}{23.325} & 0.725 & \textcolor{green}{0.286}
& 19.864 & \textcolor{cyan}{0.747} & \textcolor{green}{0.236}
& \textcolor{green}{23.462} & 0.832 & 0.128 \\
GPP-LLIE~\cite{Zhou2025GPPLLIE} & AAAI'25
& 22.339 & 0.743 & 0.281
& 20.358 & 0.602 & 0.422
& 15.222 & 0.546 & 0.443
& 23.050 & \textcolor{red}{0.862} & \textcolor{cyan}{0.081} \\
CIDNet~\cite{Yan2025HVI} & CVPR'25
& 24.975 & \textcolor{green}{0.786} & \textcolor{green}{0.194}
& \textcolor{cyan}{23.671} & 0.681 & 0.335
& 19.990 & \textcolor{green}{0.740} & \textcolor{cyan}{0.227}
& 23.808 & \textcolor{cyan}{0.857} & \textcolor{green}{0.086} \\
LersGAN~\cite{Li2025LersGAN} & JSTARS'25
& 23.652 & 0.775 & 0.267
& 20.859 & 0.653 & 0.614
& 20.110 & 0.734 & 0.257
& 21.973 & 0.796 & 0.127 \\
SPJFNet~\cite{Zhang2026SPJFNet} & AAAI'26
& 22.133 & 0.755 & 0.289
& 23.117 & \textcolor{green}{0.729} & 0.348
& \textcolor{green}{20.444} & 0.721 & 0.307
& \textcolor{red}{24.160} & \textcolor{green}{0.853} & \textcolor{red}{0.069} \\
BEM$_{\mathrm{MC}}$~\cite{Huang2026BEM} & AAAI'26
& \textcolor{cyan}{25.826} & \textcolor{cyan}{0.802} & \textcolor{cyan}{0.185}
& 21.876 & \textcolor{cyan}{0.744} & \textcolor{cyan}{0.259}
& 19.378 & 0.722 & 0.252
& 23.070 & 0.851 & 0.089 \\
\noalign{\hrule height 1pt}
\textbf{Ours} & -
& \textcolor{red}{26.527} & \textcolor{red}{0.812} & \textcolor{red}{0.170}
& \textcolor{red}{25.476} & \textcolor{red}{0.770} & \textcolor{red}{0.211}
& \textcolor{red}{21.893} & \textcolor{red}{0.787} & \textcolor{red}{0.217}
& \textcolor{cyan}{24.129} & 0.851 & 0.094 \\
\noalign{\hrule height 1.2pt}
\end{tabular}
}
\end{table*}

\textbf{1) Positive Homogeneity Bias ($B_{\mathrm{sem}}$).} 
To counteract Type II errors (semantic fragmentation and coplanar ambiguity), we leverage the illumination-invariant semantic prior $F_{\mathrm{sem}}$ extracted from DINOv3. After spatial alignment, we project $F_{\mathrm{sem}}$ into a dense semantic embedding space via a learnable transformation $W_s$, and reshape it into a token sequence $S \in \mathbb{R}^{N \times d_s}$ per window. The Positive Homogeneity Bias is formulated as the cosine similarity within this robust semantic space:
\begin{equation}
B_{\mathrm{sem}} = \frac{\mathrm{Norm}(S)\mathrm{Norm}(S)^\top}{\tau},
\end{equation}
where $\mathrm{Norm}(\cdot)$ denotes L2 normalization and $\tau$ is a learnable temperature parameter controlling the sharpness of the semantic distribution. 
\textit{Physical Insight:} Even if two spatial regions belonging to the same semantic class (e.g., continuous water bodies) exhibit vastly different degraded responses due to non-uniform illumination, their robust semantic embeddings remain highly correlated. Consequently, $B_{\mathrm{sem}}$ provides a strong positive value, forcefully anchoring the features back to the same homogeneous cluster and promoting content-consistent enhancement.

\textbf{2) Negative Heterogeneity Penalty ($B_{\mathrm{geo}}$).}
To eliminate Type I errors (cross-boundary confusion caused by severe noise), we utilize the geometry-aware feature $G$ derived from the pseudo-3D topological prior (DA3). We compress $G$ into a compact boundary descriptor via a $1\times1$ convolution $W_g$ and tokenize it into a 1D sequence $g \in \mathbb{R}^{N}$ for each window. The Negative Heterogeneity Penalty is then defined using the absolute topological difference between local tokens:
\begin{equation}
B_{\mathrm{geo}}(i,j) = -\alpha |g_i-g_j|,
\end{equation}
where $i, j \in \{1, \dots, N\}$ index the tokens within the window, and $\alpha$ is a learnable positive scale factor.
\textit{Physical Insight:} In the bird's-eye view of remote sensing, $|g_i-g_j|$ captures sudden relative elevation changes and physical truncations (e.g., the boundary between a building roof and its cast shadow). When tokens $i$ and $j$ span across a true physical boundary, this difference spikes. Due to the negative sign, $B_{\mathrm{geo}}$ imposes an overwhelming penalty on their attention affinity. This acts as an impenetrable mathematical wall, strictly prohibiting feature propagation across distinct physical structures and thereby preventing noise-induced structural blurring.

\textbf{3) Prior-Guided Aggregation.}
The final attention logits $L$ synergistically integrate the internal data-driven observation with the dual external physical constraints:
\begin{equation}
L = \frac{QK^\top}{\sqrt{d}} +\lambda_s B_{\mathrm{sem}} +\lambda_g B_{\mathrm{geo}},
\end{equation}
where $\lambda_s,\lambda_g \ge 0$ are non-negative learnable global gating scalars that adaptively balance the contributions of regional homogeneity and boundary heterogeneity based on the local degradation severity. The refined feature is then obtained through the guided aggregation:
\begin{equation}
Z^{\prime} = Z+\gamma W_O\big(\mathrm{Softmax}(L)V\big),
\end{equation}
where $W_O(\cdot)$ is the output projection (mapping sequences back to the spatial grid) and $\gamma$ is a learnable residual scale. 

Through this elegant mathematical formulation, H2CAM guarantees that feature aggregation is strictly bounded by the physical laws of the scene: tokens are strongly encouraged to aggregate within semantically homogeneous regions ($\lambda_s B_{\mathrm{sem}}$) and strictly penalized if they attempt to cross topologically heterogeneous boundaries ($\lambda_g B_{\mathrm{geo}}$). This fundamentally resolves the \textit{Attention Drift} bottleneck in extreme low-light remote sensing image enhancement.

\section{Experiments}

\subsection{Experimental Settings}

\textbf{Datasets and Benchmarks.}
To evaluate the robustness and generalization of HALO, we conduct extensive experiments across synthetic and real-world benchmarks. For synthetic low-light remote sensing enhancement, we utilize iSAID-dark and iSAID-dark (high-pixel)~\cite{Yao2024Spatial}. To rigorously assess zero-shot cross-dataset generalization, we synthesize a new benchmark, UCM-RSLL, derived from the UC Merced dataset~\cite{yang2010bag} using a unified physical degradation pipeline. We also evaluate the fundamental structural preservation capability on LOLv1~\cite{Chen2018Retinex} and LSRW-Nikon~\cite{Hai2023R2RNet}. For real-world no-reference evaluation, we employ DarkRS~\cite{Yao2024Spatial} and U3D~\cite{Lu2025U3D}. Furthermore, to demonstrate that our prior-guided aggregation preserves machine vision-oriented discriminative features, we evaluate downstream object detection performance on DOTA-v1.0~\cite{xia2018dota}. Detailed construction protocols of UCM-RSLL are provided in \textcolor{blue}{Appendix Sec.~II}.

\textbf{Evaluation Metrics.}
For synthetic benchmarks, we use PSNR, SSIM~\cite{wang2004image}, and LPIPS~\cite{Zhang2018LPIPS}. For no-reference real world datasets, we adopt BRISQUE~\cite{Mittal2012BRISQUE}, NIQE~\cite{Mittal2013NIQE}, PIQE~\cite{Venkatanath2015PIQE}, and LOE~\cite{Guo2017LIME}. For downstream object detection, we report mAP@0.5 and F1 Score to quantify the retention of discriminative features.

\textbf{Implementation Details.}
HALO is implemented in PyTorch. The semantic and geometric priors are extracted using frozen DINOv3 ViT L/16~\cite{Simeoni2025DINOv3} and Depth Anything 3~\cite{Lin2026Depth} models, respectively, demonstrating zero-shot generalization without fine-tuning. We train two separate models on LOLv1~\cite{Chen2018Retinex} and iSAID-dark~\cite{Yao2024Spatial}. The model is optimized using Adam ($\beta_1=0.9, \beta_2=0.999$) with an initial learning rate of $1\times10^{-4}$ decayed via cosine annealing. Notably, to rigorously verify cross-dataset generalization, evaluations on iSAID-dark (high-pixel), UCM-RSLL, DarkRS, and U3D are conducted directly using the model trained solely on iSAID-dark. Details of public evaluation benchmarks are provided in \textcolor{blue}{Appendix Sec.~I}, and training hyperparameters, including objective functions, are provided in \textcolor{blue}{Appendix Sec.~III}.

\begin{figure*}[tb]
\centering
\includegraphics[width=\textwidth]{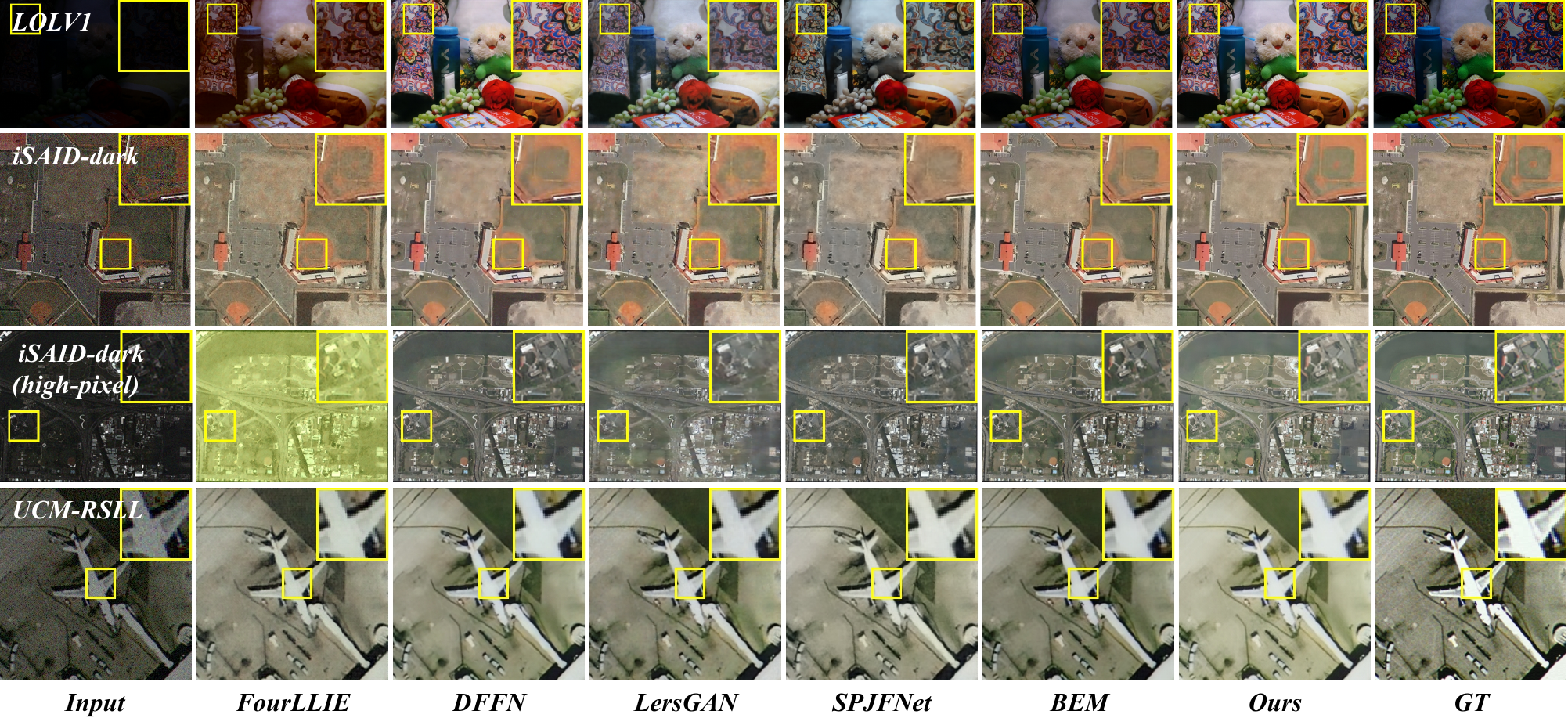}
\caption{Visual comparisons of the enhanced results by different methods on LOLv1~\cite{Chen2018Retinex}, iSAID-dark~\cite{Yao2024Spatial}, iSAID-dark (high-pixel)~\cite{Yao2024Spatial}, and UCM-RSLL. We select one representative image from each dataset. HALO restores clearer structural details and more natural illumination, while better suppressing color distortion and boundary artifacts under severe low-light degradation. \textbf{(Zoom in for the best view.)}}
\label{fig:isaid_dark}
\end{figure*}

\subsection{Low-Light Remote Sensing Image Enhancement}

\textbf{Comparisons on Synthetic Benchmarks.} We compare HALO with 14 representative methods (e.g., Uformer~\cite{Wang2022Uformer}). As shown in Table~\ref{tab:quantitative_comparison_main}, HALO achieves state-of-the-art performance on iSAID-dark, iSAID-dark (high-pixel), and UCM-RSLL. Notably, on iSAID-dark, it attains 26.527 dB, 0.812, and 0.170 in PSNR, SSIM, and LPIPS, respectively, substantially outperforming remote sensing methods like DFFN~\cite{Yao2024Spatial} and LersGAN~\cite{Li2025LersGAN}.

Beyond quantitative gains, visual comparisons (Fig.~\ref{fig:isaid_dark}) demonstrate our structural superiority. Under severe low-light degradation, unconstrained feature aggregation in existing baselines causes pronounced cross-boundary blurring and color distortion. For example, DFFN~\cite{Yao2024Spatial} and LersGAN~\cite{Li2025LersGAN} blur field boundaries in iSAID-dark and deform aircraft wings in UCM-RSLL, whereas FourLLIE~\cite{Wang2023FourLLIE} introduces global color shifts on high-resolution patches. In contrast, explicit dual-prior guidance enables HALO to effectively suppress semantic leakage, consistently reconstructing sharp physical boundaries and faithful colors across complex scenes.

Despite optimal performance on remote sensing benchmarks, HALO ranks second in PSNR (24.129 dB) on the LOLv1 dataset~\cite{Chen2018Retinex}, slightly behind  SPJFNet~\cite{Zhang2026SPJFNet}. This occurs because LOLv1 features general indoor close-ups with simple geometric structures~\cite{Chen2018Retinex}. Our dual-prior mechanism targets complex, continuously distributed Earth observation scenes prone to coplanar ambiguity and attention drift. In indoor scenarios lacking distinct semantic boundaries and pseudo-3D topological constraints, DA3 and DINOv3 guidance cannot be fully exploited. Furthermore, unlike methods fitting indoor noise distributions, our approach prioritizes cross-boundary structure preservation under extreme remote sensing degradation. Nevertheless, visual results (Fig.~\ref{fig:isaid_dark}, top row) confirm HALO is extremely robust in restoring authentic colors without noise amplification, validating its cross-domain generalization.

\begin{figure}[tb]
\centering
\includegraphics[width=\columnwidth]{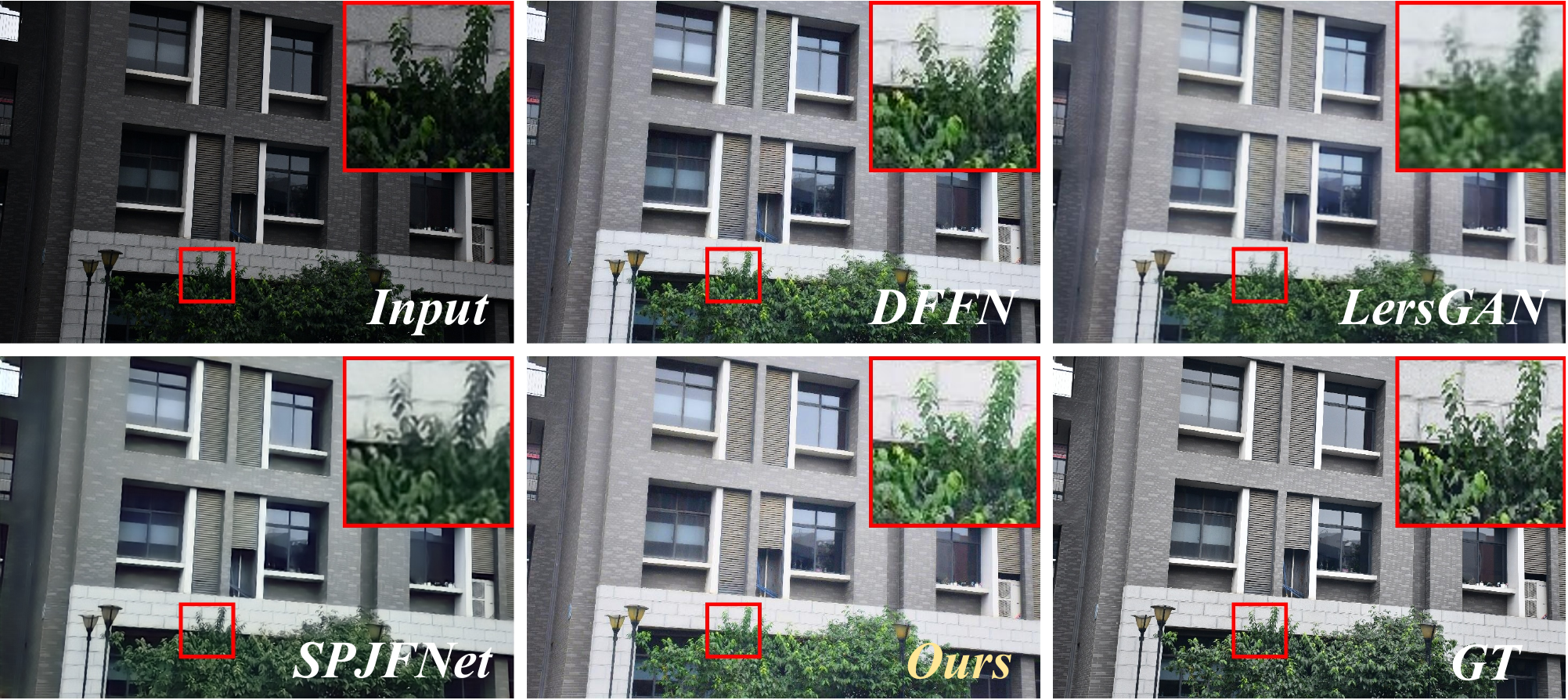}
\caption{Visual comparison on the LSRW-Nikon dataset~\cite{Hai2023R2RNet}. Compared with other methods, HALO is closer to the ground truth and better preserves wall textures, window structures, and natural illumination.}
\label{fig:lsrw}
\end{figure}

Table~\ref{tab:lsrw_nikon} reports results on the LSRW-Nikon dataset~\cite{Hai2023R2RNet}. Because most methods perform better on LSRW-Huawei~\cite{Hai2023R2RNet}, we focus on the more challenging and discriminative LSRW-Nikon subset. Here, HALO achieves superior perceptual quality and color fidelity, reflected by improved LPIPS, MAE, and $\Delta E_{00}$. Visually (Fig.~\ref{fig:lsrw}), baselines like DFFN~\cite{Yao2024Spatial} and LersGAN~\cite{Li2025LersGAN} blur fine-grained leaf details, whereas HALO accurately preserves wall textures, window structures, and natural illumination, aligning closely with the ground truth. \textcolor{blue}{\textit{Additional visual comparisons are provided in Appendix Sec.~V.}}

\textbf{Comparisons on Real-World No-Reference Benchmarks.} To evaluate HALO under complex real-world degradation, we conduct no-reference experiments on U3D~\cite{Lu2025U3D} and DarkRS~\cite{Yao2024Spatial}. Table~\ref{tab:no_reference_comparison} reports quantitative results and 4K ($3840 \times 2160$) inference times measured on U3D~\cite{Lu2025U3D} using a single Tesla V100-PCIE-32GB GPU. HALO exhibits exceptional generalization, ranking first in BRISQUE, NIQE, and LOE on U3D~\cite{Lu2025U3D}, and securing top-three results in BRISQUE, PIQE, and LOE on DarkRS~\cite{Yao2024Spatial}. 

% Although processing ultra-high-resolution images incurs more computational overhead than lightweight baselines (primarily due to explicitly extracting dual foundation-model priors DINOv3 and DA3), HALO delivers significantly superior perceptual quality. 

Although processing ultra-high-resolution images incurs a higher total computational overhead ($3966.64$ ms) compared to lightweight baselines, this latency is heavily dominated by the explicit extraction of the dual foundation-model priors. Specifically, the inference of the frozen DINOv3 and DA3 models accounts for approximately 2073.23 ms, whereas our core HALO enhancement network requires 1893.41 ms. This demonstrates that the architectural complexity of our guided aggregation mechanism remains highly competitive. Despite the offline pre-processing overhead introduced by the external priors, HALO delivers significantly superior perceptual quality. 

\begin{figure*}[tb]
\centering
\includegraphics[width=\textwidth]{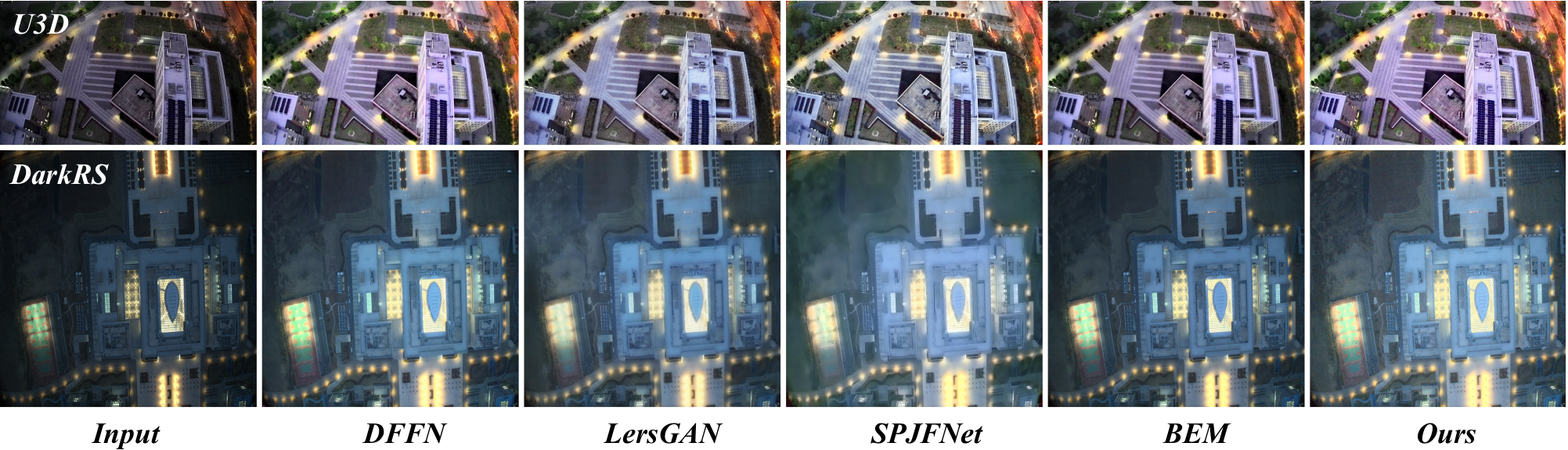}
\caption{Visual comparison of enhancement methods on U3D~\cite{Lu2025U3D} and DarkRS~\cite{Yao2024Spatial}. HALO provides significantly sharper boundaries and more faithful colors, minimizing cross-boundary color confusion and detail loss. \textbf{(Zoom in for the best view.)}}
\label{fig:darkrs}
\end{figure*}

As can be seen in Fig.~\ref{fig:darkrs}, existing baselines suffer from severe over-saturation and color leakage, including local overexposure in DFFN~\cite{Yao2024Spatial}, fluorescent distortion in LersGAN~\cite{Li2025LersGAN}, a plastic appearance in SPJFNet~\cite{Zhang2026SPJFNet}, and oversaturation in BEM~\cite{Huang2026BEM}. In contrast, HALO strictly maintains physical consistency, restoring authentic materials and natural illumination while establishing sharp physical boundaries to effectively block spectral mixing and semantic leakage in complex remote sensing scenes.

\begin{table*}[h]
\centering
\caption{Quantitative comparisons on real-world no-reference datasets, including U3D~\cite{Lu2025U3D} and DarkRS datasets~\cite{Yao2024Spatial}. The best, second-best, and third-best results are highlighted in \textcolor{red}{red}, \textcolor{cyan}{cyan}, and \textcolor{green}{green}, respectively.}
\label{tab:no_reference_comparison}
\renewcommand{\arraystretch}{1.1} 
\resizebox{\textwidth}{!}{
\begin{tabular}{l|c|c|cccc|cccc}
\noalign{\hrule height 1.2pt}
\multirow{2}{*}{Methods} & \multirow{2}{*}{Venue} & \multirow{2}{*}{Time (ms)} 
& \multicolumn{4}{c|}{U3D} 
& \multicolumn{4}{c}{DarkRS} \\
\cline{4-11}
& & & \rule{0pt}{8pt} BRISQUE$\downarrow$ & NIQE$\downarrow$ & PIQE$\downarrow$ & LOE$\downarrow$ 
& BRISQUE$\downarrow$ & NIQE$\downarrow$ & PIQE$\downarrow$ & LOE$\downarrow$ \\
\hline
Zero-DCE++~\cite{Li2022ZeroDCEPP} & TPAMI'22 & \textcolor{red}{287.35} 
& 29.149 & 3.494 & 47.701 & 0.115 
& 23.945 & 4.792 & 37.685 & 0.105 \\
FourLLIE~\cite{Wang2023FourLLIE} & ACM MM'23 & 1128.89 
& 17.735 & 3.606 & \textcolor{cyan}{27.299} & \textcolor{green}{0.051} 
& \textcolor{green}{17.411} & 5.272 & 32.106 & 0.197 \\
DFFN~\cite{Yao2024Spatial} & TGRS'24 & 797.46
& \textcolor{green}{17.256} & \textcolor{cyan}{3.466} & \textcolor{red}{24.903} & \textcolor{green}{0.051} 
& \textcolor{red}{10.085} & \textcolor{cyan}{3.618} & \textcolor{red}{25.134} & \textcolor{cyan}{0.026} \\
LersGAN~\cite{Li2025LersGAN} & JSTARS'25 & \textcolor{cyan}{472.86} 
& 17.602 & 4.040 & 34.518 & 0.132 
& 26.352 & 4.351 & 44.305 & 0.116 \\
CIDNet~\cite{Yan2025HVI} & CVPR'25 & 1639.63 
& \textcolor{cyan}{16.783} & \textcolor{green}{3.475} & \textcolor{green}{27.463} & 0.055 
& 21.157 & \textcolor{red}{3.398} & \textcolor{cyan}{26.965} & \textcolor{red}{0.018} \\
SPJFNet~\cite{Zhang2026SPJFNet} & AAAI'26 & \textcolor{green}{667.28} 
& 40.355 & 5.256 & 41.071 & \textcolor{cyan}{0.050} 
& 28.850 & 5.014 & 48.505 & 0.068 \\
BEM$_{\mathrm{MC}}$~\cite{Huang2026BEM} & AAAI'26 & 1552.99 
& 23.140 & 4.028 & 40.520 & 0.052 
& 21.543 & \textcolor{green}{4.309} & 46.067 & 0.035 \\
\noalign{\hrule height 1pt}
\textbf{Ours} & - & 3966.64 
& \textcolor{red}{16.198} & \textcolor{red}{3.342} & 27.993 & \textcolor{red}{0.049} 
& \textcolor{cyan}{12.776} & 4.909 & \textcolor{green}{27.272} & \textcolor{green}{0.029} \\
\noalign{\hrule height 1.2pt}
\end{tabular}
}
\end{table*}

\begin{table}[t]
\centering
\caption{Quantitative comparisons on the LSRW-Nikon dataset~\cite{Hai2023R2RNet} for low-light enhancement. The best and second-best results are highlighted in \textcolor{red}{red} and \textcolor{cyan}{cyan}, respectively.}
\label{tab:lsrw_nikon}
\renewcommand{\arraystretch}{1.05}
\setlength{\tabcolsep}{3pt}
\resizebox{\columnwidth}{!}{
\begin{tabular}{l|c|ccccc}
\noalign{\hrule height 1.2pt}
\multirow{2}{*}{Methods} & \multirow{2}{*}{Venue} & \multicolumn{5}{c}{LSRW-Nikon} \\
\cline{3-7}
& & \rule{0pt}{8pt} PSNR$\uparrow$ & SSIM$\uparrow$ & LPIPS$\downarrow$ & MAE$\downarrow$ & $\Delta E_{00}\downarrow$ \\
\hline
FourLLIE~\cite{Wang2023FourLLIE} & ACM MM'23 & 15.830 & 0.478 & 0.421 & 34.427 & 13.394 \\
DFFN~\cite{Yao2024Spatial} & TGRS'24 & \textcolor{red}{16.606} & 0.481 & \textcolor{cyan}{0.230} & \textcolor{cyan}{30.600} & \textcolor{cyan}{11.656} \\
GPP-LLIE~\cite{Zhou2025GPPLLIE} & AAAI'25 & 15.231 & 0.405 & 0.432 & 37.230 & 15.321 \\
LersGAN~\cite{Li2025LersGAN} & JSTARS'25 & 16.102 & \textcolor{red}{0.572} & 0.437 & 33.150 & 12.270 \\
%CIDNet~\cite{Yan2025HVI} & CVPR'25 & \textcolor{red}{16.683} & 0.468 & 0.246 & 31.304 & 12.083 \\
SPJFNet~\cite{Zhang2026SPJFNet} & AAAI'26 & 16.504 & \textcolor{cyan}{0.503} & 0.405 & 30.870 & 12.182 \\
\noalign{\hrule height 1pt}
\textbf{Ours} & - & \textcolor{cyan}{16.530} & 0.470 & \textcolor{red}{0.214} & \textcolor{red}{30.294} & \textcolor{red}{11.618} \\
\noalign{\hrule height 1.2pt}
\end{tabular}
}
\end{table}

\begin{figure*}[tb]
\centering
\includegraphics[width=0.95\textwidth]{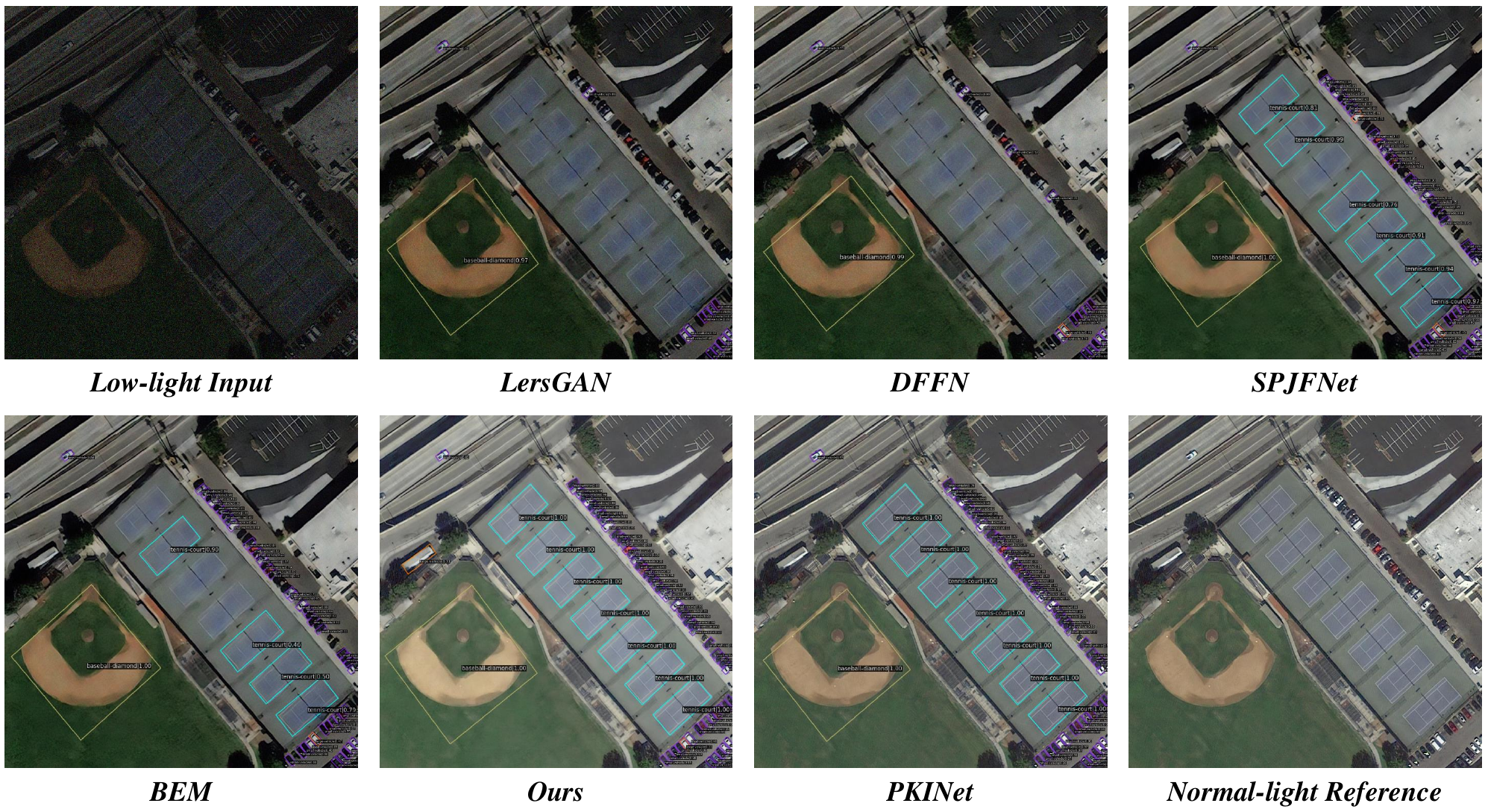}
\caption{Visual comparisons of object detection on the low-light DOTA-v1.0 dataset~\cite{xia2018dota}. By preserving sharp contours and color fidelity, our HALO avoids the semantic leakage seen in baselines, yielding significantly more accurate bounding boxes. \textbf{(Zoom in for the best view.)}}
\label{fig:detection}
\end{figure*}

\subsection{Remote Sensing Object Detection}
To further evaluate the fidelity and practical utility of our method in downstream tasks, we conduct object detection experiments on the DOTA-v1.0 dataset~\cite{xia2018dota}.
Specifically, we apply the unified low-light degradation pipeline to generate a degraded version of DOTA-v1.0. The images are enhanced using HALO and several state-of-the-art baselines, including LersGAN~\cite{Li2025LersGAN}, DFFN~\cite{Yao2024Spatial}, SPJFNet~\cite{Zhang2026SPJFNet}, and BEM$_{\mathrm{MC}}$~\cite{Huang2026BEM}, and subsequently fed into a pre-trained PKINet~\cite{cai2024poly} detector without any fine-tuning. The detection results on the original clean DOTA-v1.0 images serve as the theoretical upper bound.

As reported in Table~\ref{tab:object_detection} and Fig.~\ref{fig:detection}, existing methods suffer from significant performance degradation. Their unconstrained feature aggregation induces semantic leakage across object boundaries, leading to missing detections and inaccurate localization. Visually, baseline methods generally exhibit blurred contours and inconsistent illumination, which weakens object separability and causes both false negatives and fragmented predictions.

In contrast, HALO achieves an F1-Score of 0.790 and an mAP@0.5 of 0.732, significantly outperforming all baselines. As shown in Fig.~\ref{fig:detection}, HALO preserves sharper object boundaries and more consistent color responses, enabling more accurate and tightly aligned bounding boxes, especially in dense regions. By enforcing regional homogeneity and boundary heterogeneity at the bottleneck, HALO effectively mitigates cross-object feature mixing, thereby narrowing the gap to the clean-image upper bound and demonstrating strong effectiveness for downstream vision tasks.

\begin{table}[t]
\centering
\caption{Object detection results on the low-light DOTA-v1.0 dataset~\cite{xia2018dota}.}
\label{tab:object_detection}
\renewcommand{\arraystretch}{1.1}
\setlength{\tabcolsep}{4pt}
\resizebox{\columnwidth}{!}{
\begin{tabular}{l|c|cccc|c}
\noalign{\hrule height 1.2pt}
Method & PKINet & LersGAN & DFFN & SPJFNet & BEM & \textbf{Ours} \\
\hline
F1-Score$\uparrow$ & \textcolor{gray}{0.913}  & 0.611 & 0.635 & 0.637 & 0.719 & \textbf{0.790} \\
mAP@0.5$\uparrow$  & \textcolor{gray}{0.870}  & 0.533 & 0.546 & 0.551 & 0.640 & \textbf{0.732} \\
\noalign{\hrule height 1.2pt}
\end{tabular}
}
\end{table}

\subsection{Ablation and Robustness Study}

\textbf{Effectiveness of Dual-Prior Guidance.}
Table~\ref{tab:ablation_main} investigates the contributions of the geometric and semantic priors. Introducing either prior individually via simple feature concatenation improves upon the baseline (24.255 dB), yielding 25.971 dB for geometric guidance and 25.442 dB for semantic guidance. However, naively combining both priors without explicit mathematical coordination paradoxically degrades the PSNR to 25.315 dB. This performance drop exposes the severe \textbf{Feature Assimilation} bottleneck: under extreme degradation, simply appending prior features leads to gradient conflicts, as the unconstrained attention mechanism easily dilutes the pristine priors with corrupted internal statistics. Equipped with H2CAM, the full model resolves this by elevating the priors from auxiliary inputs to hard physical constraints, achieving the highest PSNR of 26.527 dB. This proves that explicitly coordinating semantic and geometric constraints is essential for breaking the assimilation bottleneck.

\begin{table}[t]
    \centering
    \caption{Ablation of dual-prior guidance and collaborative refinement on the iSAID-dark dataset~\cite{Yao2024Spatial}.}
    \label{tab:ablation_main}
    \renewcommand{\arraystretch}{1.1}
    \setlength{\tabcolsep}{4pt}
    \resizebox{\columnwidth}{!}{
    \begin{tabular}{lccc|ccc}
    \noalign{\hrule height 1.2pt}
    Setting & Geo. & Sem. & H2CAM & PSNR & SSIM & LPIPS \\
    \hline
    Baseline    & \xmark & \xmark & \xmark & 24.255 & 0.763 & 0.240 \\
    Geo-guided  & \checkmark & \xmark & \xmark & 25.971 & 0.799 & 0.223 \\
    Sem.-guided & \xmark & \checkmark & \xmark & 25.442 & 0.775 & 0.192 \\
    Dual-prior  & \checkmark & \checkmark & \xmark & 25.315 & 0.784 & 0.226 \\
    Full model  & \checkmark & \checkmark & \checkmark & \textbf{26.527} & \textbf{0.812} & \textbf{0.170} \\
    \noalign{\hrule height 1.2pt}
    \end{tabular}
    }
\end{table}

\begin{table}[t]
\centering
\caption{Ablation of H2CAM components on the iSAID-dark dataset~\cite{Yao2024Spatial}.}
\label{tab:ablation_h2cam}
\renewcommand{\arraystretch}{1.1}
\setlength{\tabcolsep}{4pt}
\resizebox{\columnwidth}{!}{
\begin{tabular}{lccc|ccc}
\noalign{\hrule height 1.2pt}
Variant & Local Attn. & $B_{\mathrm{sem}}$ & $B_{\mathrm{geo}}$ & PSNR & SSIM & LPIPS \\
\hline
w/o H2CAM            & \xmark     & \xmark     & \xmark     & 25.315 & 0.784 & 0.226 \\
Self-Attn only        & \checkmark & \xmark     & \xmark     & 25.621 & 0.792 & 0.211 \\
+ semantic bias only  & \checkmark & \checkmark & \xmark     & 26.145 & 0.803 & 0.188 \\
+ geometric bias only & \checkmark & \xmark     & \checkmark & 26.082 & 0.801 & 0.191 \\
Full H2CAM            & \checkmark & \checkmark & \checkmark & \textbf{26.527} & \textbf{0.812} & \textbf{0.170} \\
\noalign{\hrule height 1.2pt}
\end{tabular}
}
\end{table}

% 折线图或者柱状图
\begin{table}[t]
\centering
\caption{Robustness to controlled Gaussian noise perturbations on the UCM-RSLL dataset.}
\label{tab:robustness_noise}
\renewcommand{\arraystretch}{1.2}
\resizebox{\columnwidth}{!}{
\begin{tabular}{l|ccccc}
\noalign{\hrule height 1.2pt}
Metric & $\sigma=0$ & $\sigma=5$ & $\sigma=10$ & $\sigma=15$ & $\sigma=20$ \\
\hline
PSNR$\uparrow$ & \textbf{21.893} & 21.345 & 20.678 & 19.854 & 18.921 \\
SSIM$\uparrow$ & \textbf{0.787} & 0.765 & 0.732 & 0.695 & 0.654 \\
LPIPS$\downarrow$ & \textbf{0.217} & 0.235 & 0.262 & 0.298 & 0.341 \\
\noalign{\hrule height 1.2pt}
\end{tabular}
}
\end{table}

\textbf{Component Analysis of H2CAM.}
Table~\ref{tab:ablation_h2cam} evaluates the internal mechanics of H2CAM. Starting from the uncoordinated dual-prior baseline, relying solely on standard self-attention (Self-Attn only) provides a modest 0.306 dB improvement, as the purely data-driven affinity matrix remains highly vulnerable to \textit{Attention Drift}. Injecting the \textbf{Positive Homogeneity Bias ($B_{\mathrm{sem}}$)} substantially improves PSNR by 0.830 dB by forcefully anchoring tokens to their true semantic clusters, directly mitigating \textit{Type II errors (coplanar ambiguity and semantic fragmentation)}. Conversely, injecting the \textbf{Negative Heterogeneity Penalty ($B_{\mathrm{geo}}$)} yields a 0.767 dB gain by establishing an impenetrable mathematical wall against cross-boundary feature flow, effectively suppressing \textit{Type I errors (noise-induced structural blurring)}. Integrating both deterministic biases establishes a rigorously bounded aggregation criterion, yielding the optimal performance.

\textbf{Robustness to Severe Noise.}
To rigorously evaluate HALO's resistance to \textit{Attention Drift} under extreme conditions, Table~\ref{tab:robustness_noise} evaluates performance under varying levels of controlled Gaussian noise on the UCM-RSLL dataset. As noise increases to $\sigma=20$, attention affinities from degraded features alone may become unreliable, potentially causing cross-boundary confusion. In contrast, HALO degrades gradually, with PSNR decreasing from 21.893 dB to 18.921 dB. This resilience proves that our foundation-model priors remain inherently invariant to photometric corruption. By anchoring the feature aggregation to these robust external physical cues, HALO effectively neutralizes noise-induced structural blurring and maintains deterministic restoration trajectories even when the local SNR approaches zero.

% \section{Conclusion}
% In this paper, we propose HALO, a homogeneity-heterogeneity guided framework for low-light remote sensing image enhancement. HALO introduces semantic and geometric foundation-model priors to alleviate attention drift, coplanar ambiguity, and erroneous cross-boundary aggregation under severe low-light degradation. Through the proposed H2CAM, the two priors are injected into attention logits to guide reliable feature aggregation and preserve structural fidelity and content consistency. Extensive experiments demonstrate that HALO achieves competitive enhancement performance and improves robustness on paired, no-reference, cross-dataset, and downstream perception benchmarks.

\section{Conclusion}
In this paper, we propose HALO to resolve the Attention Drift problem in extreme low-light remote sensing image enhancement. We elevate foundation model priors from soft auxiliary inputs to deterministic mathematical constraints. Through the Homogeneity-Heterogeneity Cooperative Attention Module, we translate an illumination-invariant semantic prior into a \textit{Positive Homogeneity Bias} and a pseudo-3D topological prior into a \textit{Negative Heterogeneity Penalty}. This rigorous formulation effectively eliminates cross-boundary confusion and coplanar ambiguity, forcing the network to strictly obey physical truncations and semantic consistency at the core of feature aggregation. Extensive experiments across synthetic, real-world, and zero-shot cross-dataset benchmarks demonstrate that HALO achieves state-of-the-art visual fidelity and remarkable noise robustness. By preventing the erroneous aggregation of noise and heterogeneous materials, HALO maximally preserves discriminative features, providing a physically grounded and highly reliable foundation for downstream Earth observation tasks.

%\clearpage
\bibliographystyle{IEEEtran}
\bibliography{mybib}

@inproceedings{Chen2018Retinex,
  title={Deep Retinex Decomposition for Low-Light Enhancement},
  author={Chen, Wei and Wang, Wenjing and Yang, Wenhan and Liu, Jiaying},
  booktitle={Proceedings of the British Machine Vision Conference},
  year={2018}
}

@inproceedings{Guo2020ZeroDCE,
  title={Zero-Reference Deep Curve Estimation for Low-Light Image Enhancement},
  author={Guo, Chunle and Li, Chongyi and Guo, Jichang and Loy, Chen Change and Hou, Junhui and Kwong, Sam and Cong, Runmin},
  booktitle={Proceedings of the IEEE/CVF Conference on Computer Vision and Pattern Recognition},
  pages={1780--1789},
  year={2020}
}

@inproceedings{Wang2023Ultra,
  title={Ultra-High-Definition Low-Light Image Enhancement: A Benchmark and Transformer-Based Method},
  author={Wang, Tao and Zhang, Kaihao and Shen, Tianrun and Luo, Wenhan and Stenger, Bjorn and Lu, Tong},
  booktitle={Proceedings of the AAAI Conference on Artificial Intelligence},
  volume={37},
  number={3},
  pages={2654--2662},
  year={2023}
}

@article{Yao2024Spatial,
  title={Spatial--Frequency Dual-Domain Feature Fusion Network for Low-Light Remote Sensing Image Enhancement},
  author={Yao, Zishu and Fan, Guodong and Fan, Jinfu and Gan, Min and Chen, C. L. Philip},
  journal={IEEE Transactions on Geoscience and Remote Sensing},
  volume={62},
  pages={1--16},
  year={2024}
}

@inproceedings{Wang2023FourLLIE,
  title={{FourLLIE}: Boosting Low-Light Image Enhancement by Fourier Frequency Information},
  author={Wang, Chenxi and Wu, Hongjun and Jin, Zhi},
  booktitle={Proceedings of the ACM International Conference on Multimedia},
  pages={7459--7469},
  year={2023}
}

@article{Li2022Low,
  title={Low-Light Image and Video Enhancement Using Deep Learning: A Survey},
  author={Li, Chongyi and Guo, Chunle and Han, Linghao and Jiang, Jun and Cheng, Ming-Ming and Gu, Jinwei and Loy, Chen Change},
  journal={IEEE Transactions on Pattern Analysis and Machine Intelligence},
  volume={44},
  number={12},
  pages={9396--9416},
  year={2022}
}

@inproceedings{Ma2022Toward,
  title={Toward Fast, Flexible, and Robust Low-Light Image Enhancement},
  author={Ma, Long and Ma, Tengyu and Liu, Risheng and Fan, Xin and Luo, Zhongxuan},
  booktitle={Proceedings of the IEEE/CVF Conference on Computer Vision and Pattern Recognition},
  pages={5637--5646},
  year={2022}
}

@article{Awais2025Foundational,
  title={Foundation Models Defining a New Era in Vision: A Survey and Outlook},
  author={Awais, Muhammad and Naseer, Muzammal and Khan, Salman and Anwer, Rao Muhammad and Cholakkal, Hisham and Shah, Mubarak and Yang, Ming-Hsuan and Khan, Fahad Shahbaz},
  journal={IEEE Transactions on Pattern Analysis and Machine Intelligence},
  volume={47},
  number={4},
  pages={2245--2264},
  year={2025}
}

@article{Simeoni2025DINOv3,
  title={{DINOv3}},
  author={Sim{\'e}oni, Oriane and Vo, Huy V. and Seitzer, Maximilian and Baldassarre, Federico and Oquab, Maxime and Jose, Cijo and Khalidov, Vasil and Szafraniec, Marc and Yi, Seungeun and Ramamonjisoa, Micha{\"e}l and Massa, Francisco and Haziza, Daniel and Wehrstedt, Luca and Wang, Jianyuan and Darcet, Timoth{\'e}e and Moutakanni, Th{\'e}o and Sentana, Leonel and Roberts, Claire and Vedaldi, Andrea and Tolan, Jamie and Brandt, John and Couprie, Camille and Mairal, Julien and J{\'e}gou, Herv{\'e} and Labatut, Patrick and Bojanowski, Piotr},
  journal={arXiv preprint arXiv:2508.10104},
  year={2025}
}

@inproceedings{Lin2026Depth,
  title={Depth Anything 3: Recovering the Visual Space from Any Views},
  author={Lin, Haotong and Chen, Sili and Liew, Jun Hao and Chen, Donny Y. and Li, Zhenyu and Zhao, Yang and Peng, Sida and Guo, Hengkai and Zhou, Xiaowei and Shi, Guang and Feng, Jiashi and Kang, Bingyi},
  booktitle={Proceedings of the International Conference on Learning Representations},
  year={2026}
}

@inproceedings{Wu2023Learning,
  title={Learning Semantic-Aware Knowledge Guidance for Low-Light Image Enhancement},
  author={Wu, Yuhui and Pan, Chen and Wang, Guoqing and Yang, Yang and Wei, Jiwei and Li, Chongyi and Shen, Heng Tao},
  booktitle={Proceedings of the IEEE/CVF Conference on Computer Vision and Pattern Recognition},
  pages={1662--1671},
  year={2023}
}

@inproceedings{Zheng2022Semantic,
  title={Semantic-Guided Zero-Shot Learning for Low-Light Image/Video Enhancement},
  author={Zheng, Shen and Gupta, Gaurav},
  booktitle={Proceedings of the IEEE/CVF Winter Conference on Applications of Computer Vision Workshops},
  pages={581--590},
  year={2022}
}

@inproceedings{Nazeri2019EdgeConnect,
  title={EdgeConnect: Structure Guided Image Inpainting using Edge Prediction},
  author={Nazeri, Kamyar and Ng, Eric and Joseph, Tony and Qureshi, Faisal Z. and Ebrahimi, Mehran},
  booktitle={Proceedings of the IEEE/CVF International Conference on Computer Vision Workshops},
  pages={3265--3274},
  year={2019}
}

@inproceedings{Wang2024Depth,
  title={Depth-Aware Blind Image Decomposition for Real-World Adverse Weather Recovery},
  author={Wang, Chao and Zheng, Zhedong and Quan, Ruijie and Yang, Yi},
  booktitle={Proceedings of the European Conference on Computer Vision},
  pages={379--397},
  year={2024}
}

@article{Yu2025Multiprior,
  title={Multiprior Learning Via Neural Architecture Search for Blind Face Restoration},
  author={Yu, Yanjiang and Zhang, Puyang and Zhang, Kaihao and Luo, Wenhan and Li, Changsheng},
  journal={IEEE Transactions on Neural Networks and Learning Systems},
  volume={36},
  number={2},
  pages={3057--3070},
  year={2025}
}

@article{Cheng2017Remote,
  title={Remote Sensing Image Scene Classification: Benchmark and State of the Art},
  author={Cheng, Gong and Han, Junwei and Lu, Xiaoqiang},
  journal={Proceedings of the IEEE},
  volume={105},
  number={10},
  pages={1865--1883},
  year={2017}
}

@article{Zhao2025Atmospheric,
  title={Atmospheric Scattering Model and Non-Uniform Illumination Compensation for Low-Light Remote Sensing Image Enhancement},
  author={Zhao, Xiaohang and Huang, Liang and Li, Mingxuan and Han, Chengshan and Nie, Ting},
  journal={Remote Sensing},
  volume={17},
  number={12},
  pages={2069},
  year={2025}
}

@article{Zhu2017Deep,
  title={Deep Learning in Remote Sensing: A Comprehensive Review and List of Resources},
  author={Zhu, Xiao Xiang and Tuia, Devis and Mou, Lichao and Xia, Gui-Song and Zhang, Liangpei and Xu, Feng and Fraundorfer, Friedrich},
  journal={IEEE Geoscience and Remote Sensing Magazine},
  volume={5},
  number={4},
  pages={8--36},
  year={2017}
}

@article{Li2022ZeroDCEPP,
  title={Learning to Enhance Low-Light Image via Zero-Reference Deep Curve Estimation},
  author={Li, Chongyi and Guo, Chunle and Loy, Chen Change},
  journal={IEEE Transactions on Pattern Analysis and Machine Intelligence},
  volume={44},
  number={8},
  pages={4225--4238},
  year={2022}
}

@inproceedings{Wang2022Uformer,
  title={Uformer: A General U-Shaped Transformer for Image Restoration},
  author={Wang, Zhendong and Cun, Xiaodong and Bao, Jianmin and Zhou, Wengang and Liu, Jianzhuang and Li, Houqiang},
  booktitle={Proceedings of the IEEE/CVF Conference on Computer Vision and Pattern Recognition},
  pages={17683--17693},
  year={2022}
}

@inproceedings{Wu2022URetinexNet,
  title={URetinex-Net: Retinex-Based Deep Unfolding Network for Low-Light Image Enhancement},
  author={Wu, Wenhui and Weng, Jian and Zhang, Pingping and Wang, Xu and Yang, Wenhan and Jiang, Jianmin},
  booktitle={Proceedings of the IEEE/CVF Conference on Computer Vision and Pattern Recognition},
  pages={5891--5900},
  year={2022}
}

@inproceedings{Zheng2023CUE,
  title={Empowering Low-Light Image Enhancer through Customized Learnable Priors},
  author={Zheng, Naishan and Zhou, Man and Dong, Yanmeng and Rui, Xiangyu and Huang, Jie and Li, Chongyi and Zhao, Feng},
  booktitle={Proceedings of the IEEE/CVF International Conference on Computer Vision},
  pages={12525--12535},
  year={2023}
}

@inproceedings{Fu2023PairLIE,
  title={Learning a Simple Low-Light Image Enhancer From Paired Low-Light Instances},
  author={Fu, Zhenqi and Yang, Yan and Tu, Xiaotong and Huang, Yue and Ding, Xinghao and Ma, Kai-Kuang},
  booktitle={Proceedings of the IEEE/CVF Conference on Computer Vision and Pattern Recognition},
  pages={22252--22261},
  year={2023}
}

@inproceedings{Yang2023NeRCo,
  title={Implicit Neural Representation for Cooperative Low-Light Image Enhancement},
  author={Yang, Shuzhou and Ding, Moxuan and Wu, Yanmin and Li, Zihan and Zhang, Jian},
  booktitle={Proceedings of the IEEE/CVF International Conference on Computer Vision},
  pages={12872--12881},
  year={2023}
}

@article{Yang2023LANet,
  title={Learning to Adapt to Light},
  author={Yang, Kai-Fu and Cheng, Cheng and Zhao, Shi-Xuan and Yan, Hong-Mei and Zhang, Xian-Shi and Li, Yong-Jie},
  journal={International Journal of Computer Vision},
  volume={131},
  number={4},
  pages={1022--1041},
  year={2023}
}

@inproceedings{Zhou2025GPPLLIE,
  title={Low-Light Image Enhancement via Generative Perceptual Priors},
  author={Zhou, Han and Dong, Wei and Liu, Xiaohong and Zhang, Yulun and Zhai, Guangtao and Chen, Jun},
  booktitle={Proceedings of the AAAI Conference on Artificial Intelligence},
  volume={39},
  number={10},
  pages={10752--10760},
  year={2025}
}

@article{Li2025LersGAN,
  title={LersGAN: A GAN-Based Model for Low-Light Remote Sensing Image Enhancement},
  author={Li, Tianqi and Guo, Tiannuo and Xiang, Deliang},
  journal={IEEE Journal of Selected Topics in Applied Earth Observations and Remote Sensing},
  volume={18},
  pages={26489--26504},
  year={2025}
}

@inproceedings{Zhang2026SPJFNet,
  title={SPJFNet: Self-Mining Prior-Guided Joint Frequency Enhancement for Ultra-Efficient Dark Image Restoration},
  author={Zhang, Tongshun and Liu, Pingping and Zhang, Zijian and Zhou, Qiuzhan},
  booktitle={Proceedings of the AAAI Conference on Artificial Intelligence},
  volume={40},
  number={15},
  pages={12798--12806},
  year={2026}
}

@inproceedings{Huang2026BEM,
  title={Bayesian Neural Networks for One-to-Many Mapping in Image Enhancement},
  author={Huang, Guoxi and Yang, Qirui and Lin, Ruirui and Qi, Zipeng and Bull, David and Anantrasirichai, Nantheera},
  booktitle={Proceedings of the AAAI Conference on Artificial Intelligence},
  volume={40},
  number={7},
  pages={5004--5012},
  year={2026}
}

@article{wang2004image,
  title={Image quality assessment: from error visibility to structural similarity},
  author={Wang, Zhou and Bovik, Alan C and Sheikh, Hamid R and Simoncelli, Eero P},
  journal={IEEE Transactions on Image Processing},
  volume={13},
  number={4},
  pages={600--612},
  year={2004}
}

@inproceedings{Yan2025HVI,
  title={HVI: A New Color Space for Low-light Image Enhancement},
  author={Yan, Qingsen and Feng, Yixu and Zhang, Cheng and Pang, Guansong and Shi, Kangbiao and Wu, Peng and Dong, Wei and Sun, Jinqiu and Zhang, Yanning},
  booktitle={Proceedings of the IEEE/CVF Conference on Computer Vision and Pattern Recognition},
  pages={5678--5687},
  year={2025}
}

@inproceedings{Zhang2018LPIPS,
  title={The Unreasonable Effectiveness of Deep Features as a Perceptual Metric},
  author={Zhang, Richard and Isola, Phillip and Efros, Alexei A. and Shechtman, Eli and Wang, Oliver},
  booktitle={Proceedings of the IEEE/CVF Conference on Computer Vision and Pattern Recognition},
  pages={586--595},
  year={2018}
}

@article{Mittal2012BRISQUE,
  title={No-Reference Image Quality Assessment in the Spatial Domain},
  author={Mittal, Anish and Moorthy, Anush Krishna and Bovik, Alan C.},
  journal={IEEE Transactions on Image Processing},
  volume={21},
  number={12},
  pages={4695--4708},
  year={2012}
}

@article{Mittal2013NIQE,
  title={Making a Completely Blind Image Quality Analyzer},
  author={Mittal, Anish and Soundararajan, Rajiv and Bovik, Alan C.},
  journal={IEEE Signal Processing Letters},
  volume={20},
  number={3},
  pages={209--212},
  year={2013}
}

@article{Guo2017LIME,
  title={LIME: Low-Light Image Enhancement via Illumination Map Estimation},
  author={Guo, Xiaojie and Li, Yu and Ling, Haibin},
  journal={IEEE Transactions on Image Processing},
  volume={26},
  number={2},
  pages={982--993},
  year={2017}
}

@inproceedings{Venkatanath2015PIQE,
  title={Blind Image Quality Evaluation Using Perception Based Features},
  author={Venkatanath, N. and Praneeth, D. and Chandrasekhar, Bh. M. and Channappayya, S. S. and Medasani, S. S.},
  booktitle={Proceedings of the 21st National Conference on Communications},
  pages={1--6},
  year={2015}
}

@article{Hai2023R2RNet,
  title={R2RNet: Low-Light Image Enhancement via Real-Low to Real-Normal Network},
  author={Hai, Jiang and Xuan, Zhu and Yang, Ren and Hao, Yutong and Zou, Fengzhu and Lin, Fang and Han, Songchen},
  journal={Journal of Visual Communication and Image Representation},
  volume={90},
  pages={103712},
  year={2023}
}

@article{Lu2025U3D,
  title={Unsupervised Ultra-High-Resolution UAV Low-Light Image Enhancement: A Benchmark, Metric and Framework},
  author={Lu, Wei and Zhu, Lingyu and Chen, Si-Bao},
  journal={arXiv preprint arXiv:2509.01373},
  year={2025}
}

@inproceedings{cai2024poly,
  title={Poly Kernel Inception Network for Remote Sensing Detection},
  author={Cai, Xinhao and Lai, Qiuxia and Wang, Yuwei and Wang, Wenguan and Sun, Zeren and Yao, Yazhou},
  booktitle={Proceedings of the IEEE/CVF Conference on Computer Vision and Pattern Recognition},
  pages={27706--27716},
  year={2024}
}

@inproceedings{xia2018dota,
  title={DOTA: A large-scale dataset for object detection in aerial images},
  author={Xia, Gui-Song and Bai, Xiang and Ding, Jian and Zhu, Zhen and Belongie, Serge and Luo, Jiebo and Datcu, Mihai and Pelillo, Marcello and Zhang, Liangpei},
  booktitle={Proceedings of the IEEE/CVF Conference on Computer Vision and Pattern Recognition},
  pages={3974--3983},
  year={2018}
}

@inproceedings{Lu2025DeepSPG,
  title={DeepSPG: Exploring Deep Semantic Prior Guidance for Low-Light Image Enhancement with Multimodal Learning},
  author={Lu, Jialang and Zhao, Huayu and Zhai, Huiyu and Yang, Xingxing and Han, Shini},
  booktitle={Proceedings of the ACM International Conference on Multimedia Retrieval},
  pages={935--943},
  year={2025}
}

@article{ECADiff2025,
  title={Boosting Diffusion Networks with Deep External Context-Aware Encoders for Low-Light Image Enhancement},
  author={Tang, Pengliang and Wang, Yu and Men, Aidong},
  journal={Sensors},
  volume={25},
  number={23},
  pages={7232},
  year={2025}
}

@article{QWR2026,
  title={{QWR-Dec-Net}: A Quaternion-Wavelet Retinex Framework for Low-Light Image Enhancement with Applications to Remote Sensing},
  author={Frants, Vladimir and Agaian, Sos and Panetta, Karen and Grigoryan, Artyom},
  journal={Information},
  volume={17},
  number={1},
  pages={89},
  year={2026}
}

@article{TPGDiff2026,
  title={{TPGDiff}: Hierarchical Triple-Prior Guided Diffusion for Image Restoration},
  author={Tu, Yanjie and Yan, Qingsen and Niu, Axi and Tang, Jiacong},
  journal={arXiv preprint arXiv:2601.20306},
  year={2026}
}

@article{LMDIR2024,
  title={Training-Free Large Model Priors for Multiple-in-One Image Restoration},
  author={He, Xuanhua and Li, Lang and Wang, Yingying and Zheng, Hui and Cao, Ke and Yan, Keyu and Li, Rui and Xie, Chengjun and Zhang, Jie and Zhou, Man},
  journal={arXiv preprint arXiv:2407.13181},
  year={2024}
}

@article{Tang2025DSPFusion,
  title={{DSPFusion}: Image Fusion via Degradation and Semantic Dual-Prior Guidance},
  author={Tang, Linfeng and Li, Chunyu and Wang, Yeda and Wang, Guoqing and Yuan, Yixuan and Ma, Jiayi},
  journal={IEEE Transactions on Image Processing},
  volume={35},
  pages={6331--6345},
  year={2026}
}

@inproceedings{yang2010bag,
  title={Bag-of-visual-words and spatial extensions for land-use classification},
  author={Yang, Yi and Newsam, Shawn},
  booktitle={Proceedings of the ACM SIGSPATIAL International Conference on Advances in Geographic Information Systems},
  pages={270--279},
  year={2010}
}

@article{yin2025structure,
  title={Structure-Guided Diffusion Transformer for Low-Light Image Enhancement},
  author={Yin, Xiangchen and Yu, Zhenda and Jiang, Longtao and Gao, Xin and Sun, Xiao and Liu, Zhi and Yang, Xun},
  journal={IEEE Transactions on Multimedia},
  volume={27},
  pages={9505--9515},
  year={2025}
}

@article{Li2025SAIGFormer,
  title={SAIGFormer: A Spatially-Adaptive Illumination-Guided Network for Low-Light Image Enhancement},
  author={Li, Hanting and Zhou, Fei and Sun, Xin and Hua, Yang and Han, Jungong and Zhang, Liang-Jie},
  journal={arXiv preprint arXiv:2507.15520},
  year={2025}
}

@article{Lin2026PixIE,
  title={PixIE: Prompted Pixel-Space Low-Light Image Enhancement},
  author={Lin, Ruirui and Huang, Guoxi and Bull, David and Anantrasirichai, Nantheera},
  journal={arXiv preprint arXiv:2605.23531},
  year={2026}
}

\newpage



% \begin{IEEEbiography}[{\includegraphics[width=1in,height=1.25in,clip,keepaspectratio]{fig}}]{Michael Shell}
% \end{IEEEbiography}



% \vfill
% \clearpage
% \input{appendix}

\end{document}